\documentclass{article}

     \PassOptionsToPackage{numbers, sort&compress}{natbib}

\usepackage[preprint]{neurips_2025}

\usepackage[utf8]{inputenc} 
\usepackage[T1]{fontenc}    
\usepackage{hyperref}       
\usepackage{url}            
\usepackage{booktabs}       
\usepackage{amsfonts}       
\usepackage{nicefrac}       
\usepackage{microtype}      

\usepackage[table]{xcolor}
\usepackage{tcolorbox}
\usepackage{latexsym}
\usepackage{booktabs}  
\usepackage{multirow}  
\usepackage{siunitx}   
\usepackage{graphicx}  
\usepackage{xtab}      
\usepackage{longtable}
\usepackage{flushend}
\usepackage{balance}
\usepackage{subcaption}

\usepackage{inconsolata}
\usepackage{graphicx}
\usepackage{CJKutf8}
\usepackage{amsmath}
\usepackage{makecell}
\definecolor{darkgreen}{rgb}{0.0, 0.4, 0.0}
\definecolor{lightblue}{RGB}{220, 235, 255}  
\definecolor{lightgreen}{RGB}{220, 255, 220} 
\usepackage{enumitem}

\title{SCoRE: Benchmarking Long-Chain Reasoning in Commonsense Scenarios}

\author{
Weidong Zhan$^1$, Yue Wang$^2$, Nan Hu$^1$, Liming Xiao$^1$, Jingyuan Ma$^2$, Yuhang Qin$^1$, \\ \textbf{Zheng Li$^2$, Yixin Yang$^2$, Sirui Deng$^1$, Jinkun Ding$^1$, Qingxiu Dong$^2$, Wenhan Ma$^2$, Rui Li$^2$,} \\ \textbf{Weilin Luo$^3$, Qun Liu$^3$, Zhifang Sui$^2$}\footnotemark[1]
\\
\small{$^1$Center for Chinese Linguistics, Department of Chinese Language and Literature, Peking University}\\
\small{$^2$School of Computer Science, State Key Laboratory of Multimedia Information Processing, Peking University}\\
\small{$^3$Huawei Noah's Ark Lab, China}\\
\small{\texttt{{szf}@pku.edu.cn, zwd@pku.edu.cn}}
\\
}

\begin{document}

\maketitle
\begin{abstract}
Currently, long-chain reasoning remains a key challenge for large language models (LLMs) because natural texts lack sufficient explicit reasoning data.
However, existing benchmarks suffer from limitations such as narrow coverage, short reasoning paths, or high construction costs.
We introduce SCoRE (Scenario-based Commonsense Reasoning Evaluation), a benchmark that synthesizes multi-hop questions from scenario schemas of entities, relations, and logical rules to assess long-chain commonsense reasoning.
SCoRE contains 100k bilingual (Chinese–English) multiple-choice questions whose reasoning chains span 2–11 hops and are grouped into various difficulty levels. Each question is accompanied by fine-grained knowledge labels, explicit reasoning chains, and difficulty levels for diagnostic evaluation.
Evaluation results on cutting-edge LLMs such as o3-mini and Deepseek R1 shows that even the best model attains only 69.78\% accuracy on SCoRE (even only 47.91\% on the hard set), with errors often stemming from rare knowledge, logical inconsistency, and over-interpretation of simple questions. 
SCoRE offers a scalable, extensible framework for evaluating and diagnosing the long-chain commonsense reasoning abilities of LLMs and guiding future advances in model design and training. Our data and code can be found at
\url{https://github.com/pokerwf/KnowLogic}.
\end{abstract}

\begin{CJK*}{UTF8}{gbsn}

\section{Introduction} 
Reasoning is a high cognitive function that involves analyzing, inducting, and deducing new information based on existing knowledge\cite{DBLP:journals/corr/abs-2108-07258}. It plays a fundamental role in human intelligence. Evaluating the commonsense reasoning ability of large language models (LLMs) is crucial for the development of AI. This ability significantly influences LLMs' decision-making capabilities and is vital for advancing towards human-like intelligence in artificial general intelligence (AGI).

While commonsense knowledge and reasoning are both indispensable for human-like intelligence, current commonsense reasoning benchmarks mostly exhibit a systemic bias toward knowledge memorization over structured reasoning. 
Previous commonsense reasoning datasets typically relied on human annotation \citep{talmor2019commonsenseqa,talmor1commonsenseqa,boratko2020protoqa,geva2021didaristotleuselaptop,wei2024SimpleQA}, template rules \citep{weston2015towards,wang2023tram,parmar2024logicbench} or LLMs \citep{bai2024coig,sakai-etal-2024-mcsqa,sprague2024musr} for data generation. 
Human annotated datasets, such as CommonsenseQA~\cite{talmor2019commonsenseqa}, CommonsenseQA 2.0~\cite{talmor1commonsenseqa}, and ProtoQA~\cite{boratko2020protoqa}, guide crowdworkers to generate questions associated with given knowledge. However, the questions are mostly solvable through direct knowledge retrieval or single-step reasoning rather than multi-step reasoning, which appears to be too simple for large reasoning models. Moreover, the high cost of manual annotation makes it difficult to build large-scale datasets. Datasets based on template rules, such as bAbI~\cite{weston2015towards} and TRAM~\cite{wang2023tram}, connect commonsense knowledge with explicit logical rules, but their reasoning patterns are strictly confined to predefined templates, exhibiting limited diversity and short reasoning chains. Though some LLM-generated datasets, such as MuSR~\cite{sprague2024musr}, can combine high-density knowledge with multi-step reasoning, their data quality are not well guaranteed.
Furthermore, these benchmarks lack fine-grained features, which hinders a detailed analysis of model performance, and the data generation process is difficult to precisely control.

To highlight these issues, we introduce \textbf{SCoRE}, a diagnostic benchmark designed to evaluate complex logical reasoning in commonsense scenarios through knowledge-driven synthetic data generation. SCoRE contains 100,000 bilingual (Chinese and English) questions focusing on four domains of commonsense closely related to everyday human life: space, time, social, nature, along with a mix domain for questions with multiple "domain" labels with adjustable reasoning chains up to 11 hops, emphasizing the importance of complex reasoning. Table \ref{tab:commonsense} shows the cases of commonsense reasoning in scenarios of different domains.
The questions are generated by a knowledge-driven synthetic data strategy, which involves manually creating a reliable knowledge base that integrates diverse commonsense knowledge and scenarios to ensure data quality, and designing an automatic inference engine capable of controlling the entire inference process and recording fine-grained features, to generate questions and answers. By controlling features such as the length of the reasoning chain and the complexity levels of knowledge, the data can be generated at varying difficulty levels. This strategy can ensure the scale and accuracy of data with minimal labor costs. Table~\ref{tab:comparison} compares three kinds of previous datasets with ours, \textbf{SCoRE}, generated by knowledge-driven synthetic data strategy.

\begin{table*}[h!] 
  \centering
  \caption{Four domains of commonsense and reasoning cases.} 
    \footnotesize
    \resizebox{\columnwidth}{!}{
  \begin{tabular}{l | p{16cm} } 
    \toprule 
    \textbf{Domain} & \textbf{Cases of Commonsense Reasoning} \\ 
    \midrule 
    \multirow{4}{*}{Space} & \underline{David}, \underline{Jennifer}, \underline{John} and \underline{James} came to a hot-pot restaurant for a meal. They sat in a \textcolor{red}{\textbf{four-person booth}}. \underline{Two people} sat in each booth, \textcolor{red}{\textbf{facing each other}}. \underline{David} is to the \textcolor{red}{\textbf{right}} of \underline{Jennifer} on the same booth. \underline{John} is the \textcolor{red}{\textbf{right neighbour}} of \underline{James}.\\
    & Q: Who is \textcolor{red}{\textbf{diagonally opposite}} \underline{John}?\\
    & A: \underline{David}\\ \midrule
     \multirow{5}{*}{Time} & Jack is a college student, and here are his weekly plans. Jack \underline{learns Japanese} on \textcolor{blue}{\textbf{Monday}} and plays badminton \textcolor{blue}{\textbf{on Wednesday}}. \textcolor{blue}{\textbf{2 days after}} \underline{learning Japanese}, Jack has a \underline{group meeting}. \textcolor{blue}{\textbf{1 day after}} the \underline{group meeting}, Jack \underline{cleans his dormitory room}.\\
& Q: What will Jack do \textcolor{blue}{\textbf{4 days after}} he \underline{cleans his dormitory room}?\\
& A: \underline{Learning Japanese}\\ \midrule
     
\multirow{4}{*}{\makecell[l]{Social}} & \underline{Alice} is \underline{Bob}'s \textcolor{orange}{\textbf{ex-wife}}, as well as \underline{Carol}'s \textcolor{orange}{\textbf{ex-girlfriend}}. \underline{Dave} is \underline{Alice}'s \textcolor{orange}{\textbf{boss}}, a \textcolor{orange}{\textbf{friend}} of \underline{Bob} and also the \textcolor{orange}{\textbf{husband}} of \underline{Eve}, who is a \textcolor{orange}{\textbf{classmate}} of \underline{Carol}. \\
& Q: What is the relationship between \underline{Dave}'s \textcolor{orange}{\textbf{wife}} and \underline{Alice}'s \textcolor{orange}{\textbf{ex-boyfriend}}?\\
& A: \textcolor{orange}{\textbf{Classmate}}\\ \midrule
     \multirow{4}{*}{\makecell[l]{Nature}} & The four enclosures in the zoo keep \underline{carp}, \underline{duck}, \underline{turkey}, and \underline{fox}. The animal in enclosure \textcolor{darkgreen}{\textbf{No.3}} has\textcolor{darkgreen}{\textbf{ 4 less legs}} than the animal in enclosure \textcolor{darkgreen}{\textbf{No.4}}. The animal in enclosure \textcolor{darkgreen}{\textbf{No.2}} \textcolor{darkgreen}{\textbf{can swim}}.\\
    & Q: What is kept in enclosure \textcolor{darkgreen}{\textbf{No.1}} ?\\
    & A: \underline{Turkey}\\
    \bottomrule 
  \end{tabular}
  }
  \label{tab:commonsense} 
\end{table*}

\begin{table*}[ht!] 
  \centering
  \caption{Comparison of commonsense reasoning datasets}
  \footnotesize
  \resizebox{\columnwidth}{!}{
  \begin{tabular}{l | l| c | c | c | c| c}
    \toprule
    \textbf{Datasets} & \textbf{Strategy} & \textbf{\makecell{Question \\Diversity}} & \textbf{\makecell{Accuracy \\Assurance}} & \textbf{\makecell{Automated \\Generation}} & \textbf{\makecell{Long-chain \\ Reasoning}} & \textbf{\makecell{Fine-grained \\Features}} \\
    \midrule
    CommonSenseQA \citep{talmor2019commonsenseqa} & human annotation &√&√ & × & × & × \\ 
    
    TRAM \citep{wang2023tram} & template rules &×& √ & √ & × & × \\
    
    MuSR \citep{sprague2024musr} &  LLM generation&√ &  × &  √ &  √ &  × \\
    \midrule
    \textbf{SCoRE (ours)} & \makecell[l]{knowledge-driven} &√& √ & √ & √ & √ \\
    \bottomrule
  \end{tabular}
  }
  \label{tab:comparison}
\end{table*}

The contributions of this paper are as follows:

\begin{enumerate}

    \item We propose SCoRE, a bilingual benchmark specifically designed to evaluate complex logical reasoning in commonsense scenarios through multi-step inference chains. It consists of 100,000 commonsense reasoning questions integrating diverse commonsense knowledge, multiple scenarios and long reasoning chains in different difficulty levels.

     \item We propose a knowledge-driven data synthesis method combining manually construction of a small-scaled knowledge base and automatic question generation to ensure data quality and quantity at a minimal labor expense. This method has a transparent and traceable workflow that facilitates interpretable evaluations of LLMs.

    \item We conducted evaluations using 13 state-of-the-art reasoning LLMs. The highest-performing model achieved a score of 69.78\%, and the average accuracy on the hard set was only 47.91\%, indicating that SCoRE is challenging for the top-performing models.
 
    \item Through case analysis, we identified several significant shortcomings in the commonsense reasoning capabilities of LLMs, including misunderstandings of low-frequency commonsense knowledge, self-contrast in logic, and overthinking, among others.
\end{enumerate}

\section{Data Synthesis for SCoRE} 
The basic methodology for generating our reasoning data revolves constructing knowledge base and ultilizing Inference Engine to generate questions. The core knowledge base is relatively small in scale, and to ensure accuracy, it primarily leverages existing knowledge bases combined with manual construction. The reasoning engine is developed under strict logical rules, which—given the accuracy of the knowledge base—guarantees the precision of the automatically synthesized data.

\subsection{Scenario-Centered Knowledge Base Construction} 

\begin{figure}[h]
  \centering
  \includegraphics[width=1.0\linewidth]{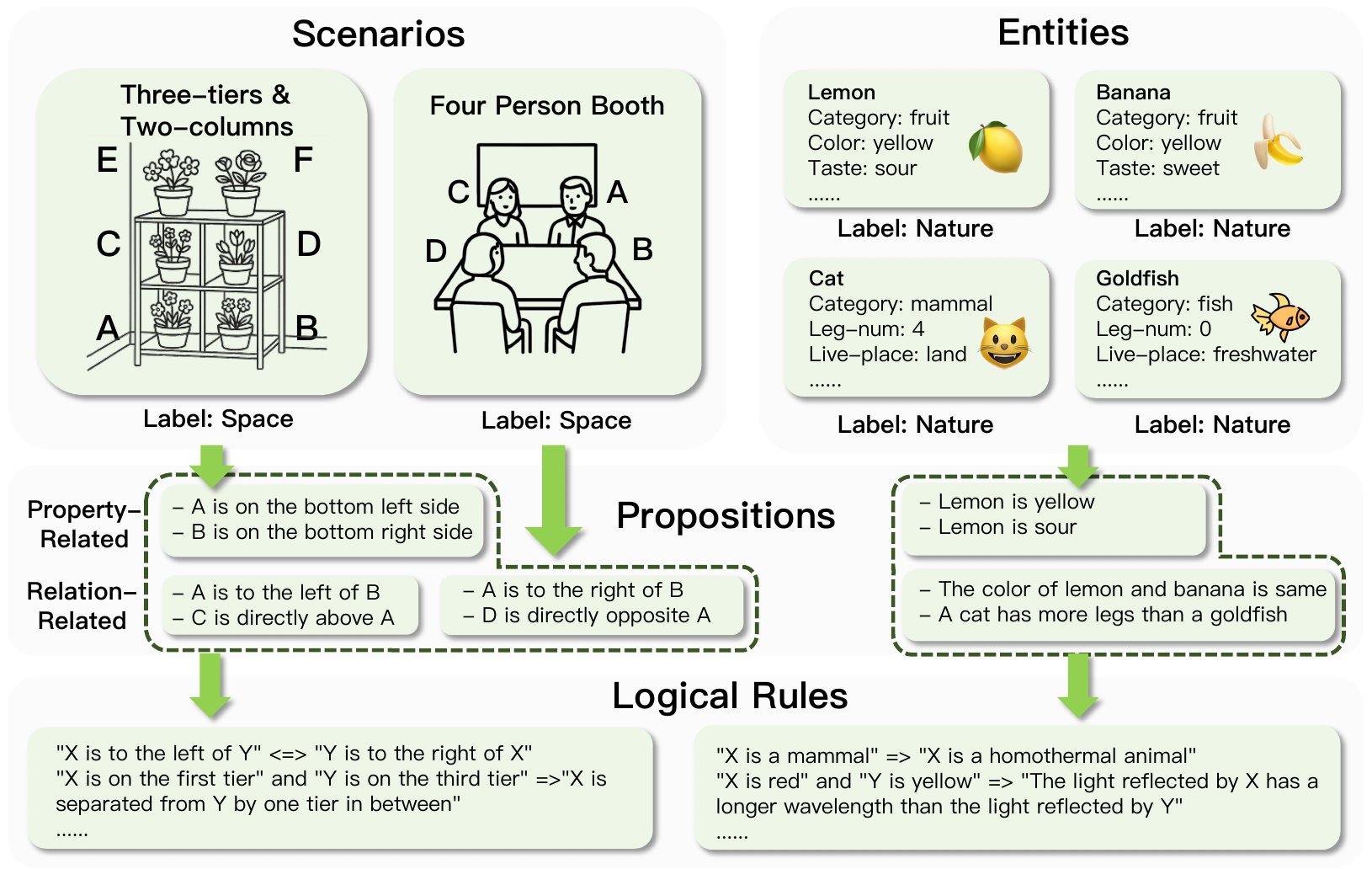}
  \caption {The framework of knowledge base. Illustrated with scenarios in the space domain and entities in the nature domain.}
  \label{fig:framework}
\end{figure}

SCoRE relies on a knowledge base that includes scenarios, entities, propositions to express properties and relations, and logical rules for reasoning. The overall framework of the knowledge base is presented in Figure \ref{fig:framework}.

Scenarios are the core of our knowledge base. We believe that commonsense reasoning requires an understanding of the specific scenarios in which additional implied information must be inferred to fully grasp the context. Each scenario has designated slots for placing entities, and these slots impose selection constraints on entities. For example, in a "four-person booth" scenario, there should be four entities, each should be a person. The scenario also constrains the way we describe the relations of entities, for example, when describing the spatial relations in a "three-tier and two-column" scenario, we may use left, right, above and below, but we would not describe the front and back of entities. Scenarios are constructed by manually designing a minimum set of  propositions based on the properties and basic relations, and a set of logical rules to infer all relations. These propositions and rules can be reused when constructing a new scenario by judging whether they are appplicable in the new scenario. Appendix \ref{app:scenario_diagram} lists the scenarios used by SCoRE.

Entities are candidates to be filled into the scenarios, which also have multiple properties and relations described by propositions and logical rules. Each scenario and entitiy is annotated with a "domain" label, and each proposition is labelled with its domains, related entities and relations for fine-grained analysis. The entities and their properties are extracted from existing external resources such as HowNet~\cite{dong2003hownet}, ConceptNet~\cite{speer2017conceptnet} and Wiki. Based on them, we manually summarize logical rules associated with the properties or relationships of the entities.

The current version of the knowledge base organizes entities, relations between entities, and factual propositions into structured scenarios representing core knowledge across spatial, temporal, natural, and social domains. At this stage, the knowledge base includes 11 scenarios, 707 entities, and 939 rules associated with 29 properties and 109 relations. It is designed to be extensible, and will be continuously expanded in future iterations.
Details about its construction are shown in appendix~\ref{app:knowledgebase}.

Though the knowledge base is relatively small in scale, it has the potential to generate a large number of questions. Taking the "three-tiers two-columns" scenario as an example, 633 entities in the knowledge base meet the constraints of this scenario. Just selecting 6 entities to fill into this scenario yields $A^6_{633} = 6*10^{16}$ possible combinations.

\begin{figure*}[h]
  \centering
  \includegraphics[width=1\linewidth]{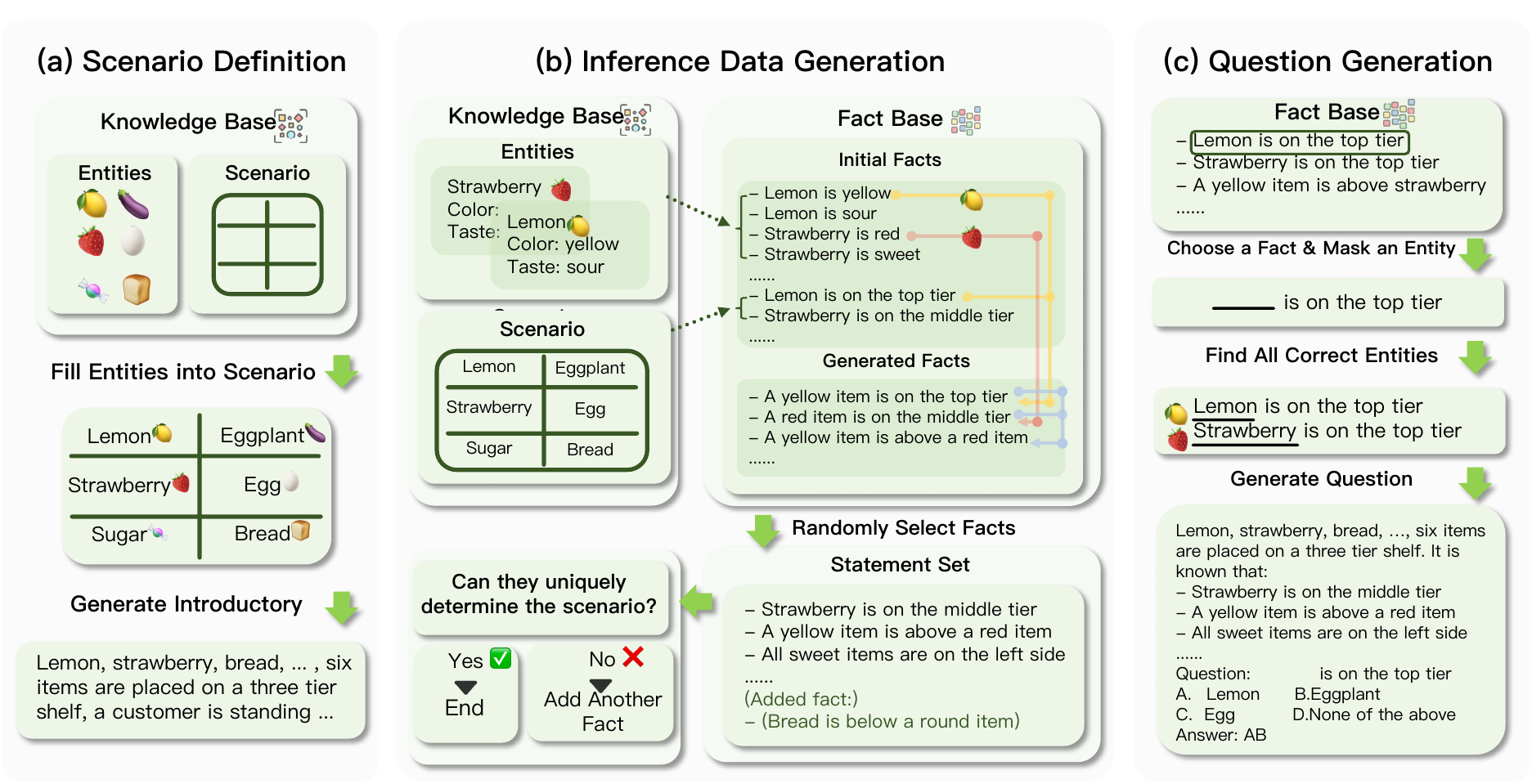}
  \caption{
Overall process of data synthesis. The pipeline consists of three stages: \textbf{(1) Scenario Definition:} selecting entities/events from the knowledge base and constructing natural language descriptions; \textbf{(2) Inference Data Generation:} applying the Reasoner to iteratively generate a fact base via rule-based logic and select a minimal statement set that uniquely determines all entity positions; \textbf{(3) Question Design:} using the statement set and answer key to generate different question types (e.g., precise, vague, true/false) with appropriate options. This process ensures verifiable reasoning chains, rich logical structures, and natural language fluency.
}
  \label{fig:data_synthesis}
\end{figure*}

\subsection{Knowledge-Driven Question generation}
%

Once the knowledge base is completed, we can ultilize a knowledge-driven method to automatically generate numerous questions.
Our Question generation is fully automated by the Inference Engine, which consists of three steps. The overall process of question generation is shown in Figure~\ref{fig:data_synthesis}.

\paragraph{Step 1: Scenario Definition}
This first step establishes the scenario and enhances its presentation. It involves selecting an scenario and  a few entities that meets the constrains of the scenario from the knowledge base and filling the entities into the scenario. Crucially, it generates introductory text for context and applies templates for natural language transformation of scenario elements. This combines scenario creation with immediate linguistic refinement for user-friendliness.

\paragraph{Step 2: Inference Data Generation}
The second step is utilising a Reasoner to generate inference data. The Reasoner is a rule-based program that takes the relations associated with the scenario and all properties of the entities or events as input. 
It maintains a fact base, initialized by the descriptions of all properties of all entities or events, such as 'strawberry is on the middle floor', 'lemon is on the top floor', 'lemon is sour', and 'strawberry is red'. 
The program then automatically traverses the logic rules in the knowledge base, matches them with the initial facts, generates new facts, and adds them to the fact base.
For example, from facts 'strawberry is on the middle floor' and 'strawberry is red' and the syllogism rule, it can generate a new fact 'a red item is on the middle floor'.
The program then takes the newly added facts in the fact base, along with the original facts, as the new initial facts and inputs them into the Reasoner again. 
This process is repeated until no new facts can be generated. Each fact is labeled with the properties or relations involved to enable fine-grained analysis.

With the fact base completed, the Reasoner selects a set of facts that can uniquely determine the slot of each entity or event in the scenario step by step.
For each step, the Reasoner randomly selects a fact from the fact base, adds it to the statement set, and verifies whether the statement set can uniquely determine the slot of each entity or event. During this process, the program automatically records the domains, entities, properties and relations involved in the statements and the number of inference steps. This process is repeated until the answer is 'Yes'. An explicit long reasoning chain can be formed through this repeated stacking of fact base generation and statement selection.

\paragraph{Step 3: Question Design}
The third step is utilising a Question Generator to design the question. The Question Generator takes the statement set and the ground-truth arrangement of entities or events as input. It first chooses a question type. If the question type is 'Correct Statement' or 'Incorrect Statement', the generator will randomly select four pairs of entities or events to produce statements, either correct or incorrect, as options. 
When the question type is 'Precise', the generator generates a proposition that can uniquely determine the slot of the entity or event in the scenario and mask the relevant information.  
When the question type is 'Vague', the generator generates a proposition that multiple entities can satisfy this proposition and identifies all those entities as potential answers.
For options, if the scenario has 4 slots, then all entities/events/slots will be the options. Otherwise, it randomly selects three of them as options A, B, and C, then add 'None of the above' as option D.


We conducted three rounds of human verification for the quality of data and the accuracy of the data reach 100\% after this process. Details of human verifications and refinement are shown in Appendix~\ref{app:human_verify}.


This approach automatically turns a small commonsense knowledge base into a large dataset of multi-step reasoning problems carrying explicit, verifiable inference chain. It offers four key benefits: (1) logical reliability—every question links back to a formally checked proof; (2) combinatorial richness—flexible rule composition yields diverse commonsense patterns; (3) controllable depth—recorded proof traces let us tune chain length and difficulty; and (4) textual naturalness—scenario grounding produces fluent, context-rich problem statements.

\section{Data Statistics} 
The full dataset of SCoRE contains 100,000 questions generated by the above-mentioned method. Considering the testing cost, we intentionally selected 6,000 questions that can cover all knowledge points in our knowledge base to form the SCoRE test dataset.




\paragraph{Knowledge Density}
SCoRE test set consists of 6,000 questions bilingual in Chinese and English spanning four major commonsense domains: space, time, social, and nature, along with a mix domain for questions with multiple "domain" labels, with 600 questions per domain per language. 
 The questions cover all knowledge points in the knowledge base, totaling 11 scenarios, 707 entities, 29 properties and 109 relations.

\paragraph{Explicit Reasoning Chains}
All facts in the question are generated by the inference engine and all reasoning steps are recorded during the data synthesis process. Due to the explicit and traceable reasoning chains in our data synthesis process, the length of these chains is highly controllable. Our dataset covers questions with reasoning chains ranging from 2 to 11 hops, enabling hierarchical evaluation of models’ multi-hop reasoning capabilities.

\paragraph{Fine-grained Labels}
Questions in SCoRE are generated by connecting domain knowledge via logical rules, ensuring a clear and traceable reasoning workflow. 7 kinds of labels are recorded during the reasoning process: domains, scenarios, entities, properties and relations used in the reasoning process, along with the reasoning chain length and question type. Questions that models get wrong can be analyzed and traced back to specific knowledge points or reasoning steps. Detailed cases are shown in Appendix \ref{app:case_study}.

\paragraph{Difficulty Levels}
Calculated based on the cognitive load, knowledge complexity and information completeness of the questions, SCoRE includes three difficulty levels: easy, medium and hard. The difficulty calculation formulas for each domain are given in Appendix \ref{app:level}. The distribution ratio of difficulty levels is 1:2:3. We believe that current LLMs are in the phase of transitioning from easy to complex reasoning tasks. Therefore, SCoRE focuses on more medium and hard questions, in order to accelerate the improvement of the model's reasoning capabilities.

\section{Experiment}
\subsection{Experiment Setup}
We evaluate a comprehensive set of existing models, including both open-source and closed-source ones. The tested models include \texttt{GPT-4o}~\citep{gpt4o}, \texttt{O1-Preview}, \texttt{O1-Mini}, \texttt{O3-Mini}, \texttt{GLM-4-Plus}, \texttt{GLM-Zero-Preview}~\citep{glm2024chatglm}, \texttt{DeepSeek-Chat (DeepSeek-V3)}~\citep{deepseekai2024deepseekv3technicalreport}, \texttt{DeepSeek-Reasoner (DeepSeek-R1)}~\citep{deepseekai2025deepseekr1incentivizingreasoningcapability}, \texttt{Qwen-Max},  \texttt{Qwen2.5-72B-Instruct}, and \texttt{QwQ-32B-Preview}~\citep{qwen2.5}. We also try the distilled model by Deepseek-R1, which is \texttt{r1-distill-qwen-32b}.~\citep{deepseekai2025deepseekr1incentivizingreasoningcapability}
The experiments are conducted in both Chinese and English, with detailed test procedures provided in Appendix~\ref{app:experiments setting}.
The answers are extracted from the model response using a rule based method. Details about the unextracted rates are shown in Appendix~\ref{app:answer_extractor}.

For human evaluation, we recruited 10 undergraduate participants for each domain and randomly sampled 5\% of the dataset as test questions, spanning all difficulty levels, scenarios and question types. Each participant was required to complete 60 questions, including 30 in Chinese and 30 in English, in a time limit of 8 hours. Details of the human evaluation are provided in Appendix \ref{app:human_evaluation}.

\subsection{Experiment Result}
The complete results are presented in Table~\ref{tab:model_performance}. Among the models tested, O1-Preview achieves the highest performance. In contrast, DeepSeek-R1, the top-performing open-source model, underperforms relative to the closed-source models. Overall, our benchmark remains a challenging test, effectively highlighting the limitations of current models across various domains. Notably, LLMs specifically trained for inference tend to outperform general LLMs within the same family when tackling such complex questions.

Compared to human performance, the maximum scores achieved by the models in the space, nature and mixed domains are significantly lower than the mean scores of human participants. The overall performance of the models also falls short of human-level accuracy, indicating a substantial gap in commonsense reasoning capabilities between LLMs and humans. However, in the time and social domain, the best performing model exceeds the mean human performance. We attribute it to the ambiguity of temporal expressions in natural language, which can lead to divergent interpretations and reasoning outcomes among human participants, and to the complexity kinship networks of the questions in hard level that exceeds the cognitive load most humans can manage. It should be noted that some human errors appear to stem from carelessness. Given sufficient time and attention, human participants would be capable of achieving significantly higher scores, potentially even perfect accuracy. For a detailed analysis, please refer to Appendix \ref{app:human_evaluation}.

\begin{table*}[htbp]
    \centering
    \caption{Models Performance on Different Domains (Accuracy \%). Best performers in each domain are \textbf{bolded}.}
    \resizebox{\textwidth}{!}{
    \begin{tabular}{l c c c c c c c c c c c}
        \toprule
        \multirow{2}{*}{\textbf{Model}} & 
        \multicolumn{2}{c}{\textbf{Space}} & 
        \multicolumn{2}{c}{\textbf{Nature}} & 
        \multicolumn{2}{c}{\textbf{Time}} & 
        \multicolumn{2}{c}{\textbf{Social}} & 
        \multicolumn{2}{c}{\textbf{Mix}} & 
        \textbf{Avg} \\
        \cmidrule(lr){2-3} \cmidrule(lr){4-5} \cmidrule(lr){6-7} \cmidrule(lr){8-9} \cmidrule(lr){10-11}
        & CN & EN & CN & EN & CN & EN & CN & EN & CN & EN & \\
        \midrule
        \multicolumn{11}{c}{\textbf{Closed-Source Models}} \\ 
        \midrule
          \rowcolor{green!5} o1-preview                 & \textbf{67.17} & \textbf{58.83} & \textbf{89.83} & \textbf{84.33} & 79.67 & 80.67 & 49.67 & 71.67 &\textbf{ 61.17} & \textbf{54.83} & \textbf{69.78} \\
          \rowcolor{green!5} o1-mini                        & 62.00 & 56.50 & 82.00 & 75.33 & \textbf{88.00} & \textbf{85.67} & 33.83 & 56.67 & 48.00 & 45.33 & 63.33 \\
          \rowcolor{green!5} claude-3.5-sonnet      & 45.33 & 43.17 & 76.00 & 70.83 & 60.50 & 71.67 & 56.83 & 68.67 & 36.33 & 36.17 & 56.55 \\
          \rowcolor{green!5} glm-zero-preview    & 39.50 & 34.50 & 73.17 & 71.83 & 66.83 & 79.17 & 38.67 & 78.33 & 22.83 & 27.33 & 53.22 \\
          \rowcolor{green!5} glm-4-plus                  & 30.33 & 30.17 & 74.17 & 64.50 & 71.67 & 70.67 & 38.67 & 58.17 & 25.50 & 26.00 & 48.98 \\
          
          \rowcolor{green!5} gpt-4o                         & 29.50 & 29.00 & 68.17 & 65.67 & 65.50 & 69.83 & 25.17 & 45.00  & 23.67 & 26.83 & 44.83 \\
        \rowcolor{green!5} qwen-max                   & 15.83 & 13.17 & 65.83 & 64.00 & 54.17 & 71.67 & 47.67 & 56.33 & 24.83 & 26.00 & 44.50 \\  
          \rowcolor{green!5} o3-mini                      & 19.67 & 22.00 & 57.33 & 57.83 & 58.50 & 62.33 & 24.67 & 70.00 & 26.67 & 25.17 & 42.41 \\
        \midrule
        \multicolumn{11}{c}{\textbf{Open-Source Models}} \\ 
        \midrule
           \rowcolor{blue!5} deepseek-r1                & 54.67 & 52.17 & 81.33 & 75.33 & 57.67 & 71.83 & \textbf{80.83} & \textbf{84.00} & 48.67 & 39.17 & 64.65 \\
          \rowcolor{blue!5} qwq-32B-preview       & 40.50 & 43.50 & 76.33 & 75.00 & 59.67 & 76.83 & 55.67 & 77.50 & 27.50 & 32.67 & 56.52\\
           \rowcolor{blue!5} deepseek-v3                 & 36.83 & 35.67 & 67.83 & 64.00 & 60.83 & 65.00 & 45.00 & 82.17 & 24.83 & 33.67 & 50.56 \\
\rowcolor{blue!5} deepseek-r1-distill-qwen-32b & 45.17 & 37.50 & 81.83 & 60.17 & 66.00 & 65.33 & 54.33 & 42.83 & 26.50 & 23.00 & 50.27\\
          \rowcolor{blue!5} qwen-2.5-72B               & 30.67 & 26.00 & 68.83 & 60.83 & 66.00 & 77.00 & 29.33 & 51.17 & 22.50 & 23.17 & 45.55\\
        \midrule
        \multicolumn{11}{c}{\textbf{Human}} \\ 
        \midrule
          \rowcolor{red!5} Best & 96.67 & 90.00 & 100.00 & 96.67 & 93.33 & 93.33 & 80.00 & 100.00 & 96.67 & 86.67 & 93.33\\
          \rowcolor{red!5} Mean & 85.71 & 85.24 & 95.19 & 94.07 & 80.42 & 80.83& 70.83 & 96.11 & 80.48 & 80.00 & 82.36 \\
        \bottomrule
    \end{tabular}
    }
    \label{tab:model_performance}
\end{table*}

\section{Analysis} 
\subsection{General Analysis}
\paragraph{More Tokens Needed for Harder Questions}
Intuitively, for reasoning models, more challenging problems typically require longer reasoning chains, resulting in increased token counts in model outputs. We conducted experiments on the relationship between token count and question difficulty. The results reveal that reasoning models exhibit a stronger correlation between question difficulty and reasoning length compared to standard models.
Details are shown in Appendix~\ref{app:token_count}.


        
        
    
\begin{figure}
    \centering
    \includegraphics[width=1\linewidth]{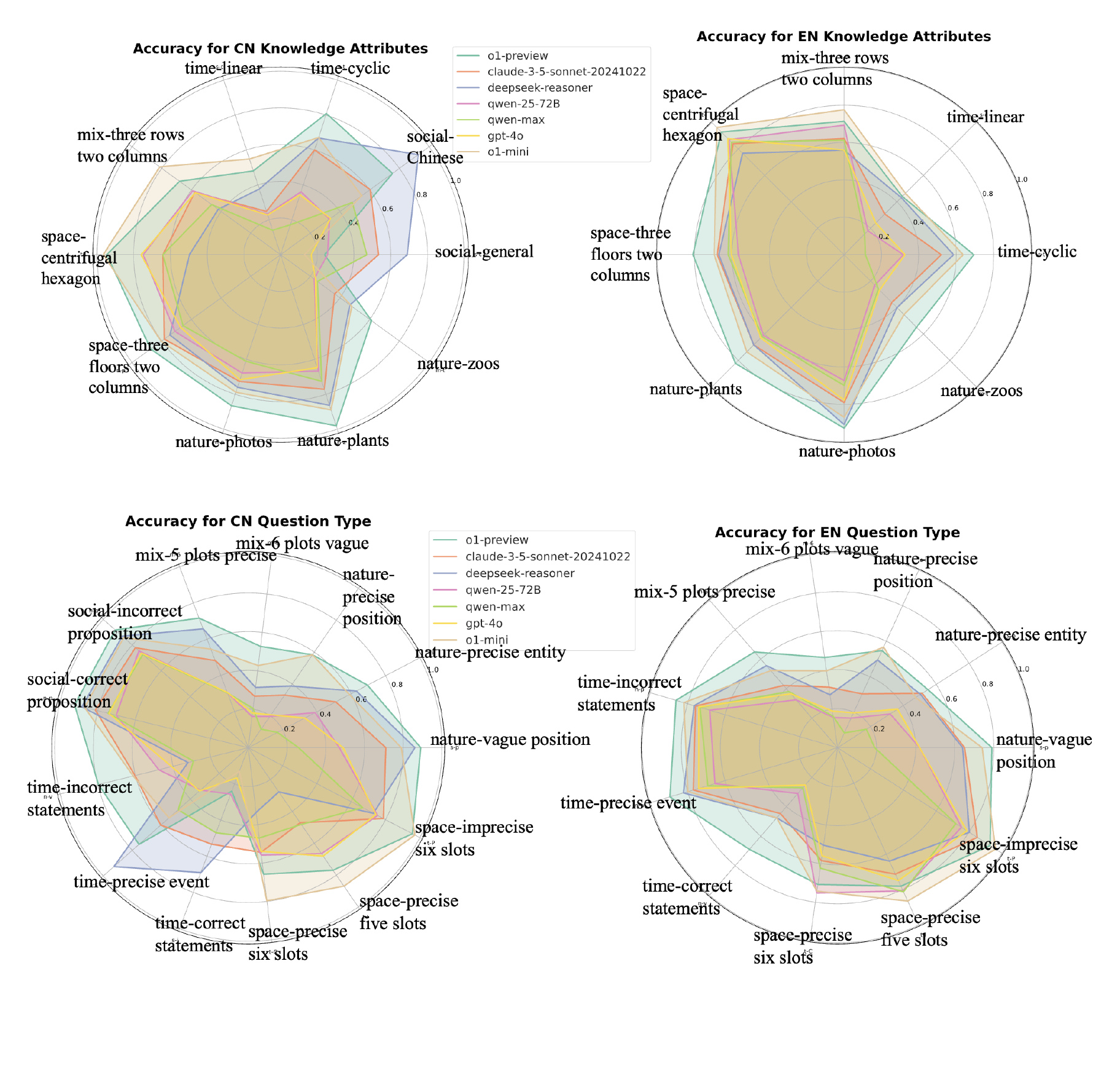}
    \caption{The performance of LLMs on Chinese and English questions with different knowledge attributes and question types. Here, if the answer to a question involves only a single entity, it is termed "precise." If it involves multiple entities, it is termed "vague".}
    \label{fig:radar}
\end{figure}

\paragraph{Lower Performance on Rarer Scenarios}
We simultaneously analyze the performance of different models across various knowledge domains and scenarios, and the results are presented in Figure~\ref{fig:radar}. The overall distribution of correct rates follows a similar trend. However, even within the same domain, there are significant performance variations across different scenarios. For instance, models perform better on general social relationships than on family relationships. Similarly, models show weaker performance when addressing spatial scenarios such as the layout of centrifugal hexagon, compared to the arrangement of three rows and two columns.


\paragraph{Changing the Way to Ask Affects Performance}
Additionally, the formulation of the question also plays a crucial role in the model’s performance. For example, as shown in Figure~\ref{fig:radar}, the phrasing of a question, such as asking the model to judge whether a statement is correct or incorrect, can influence the accuracy of the response. Similarly, the level of precision in describing spatial relations, such as the difference between vague and precise type of the six-slot scenario in space domain also affects the models to understand and respond correctly.

\begin{figure}[h!]
    \centering
    \includegraphics[width=0.8\linewidth]{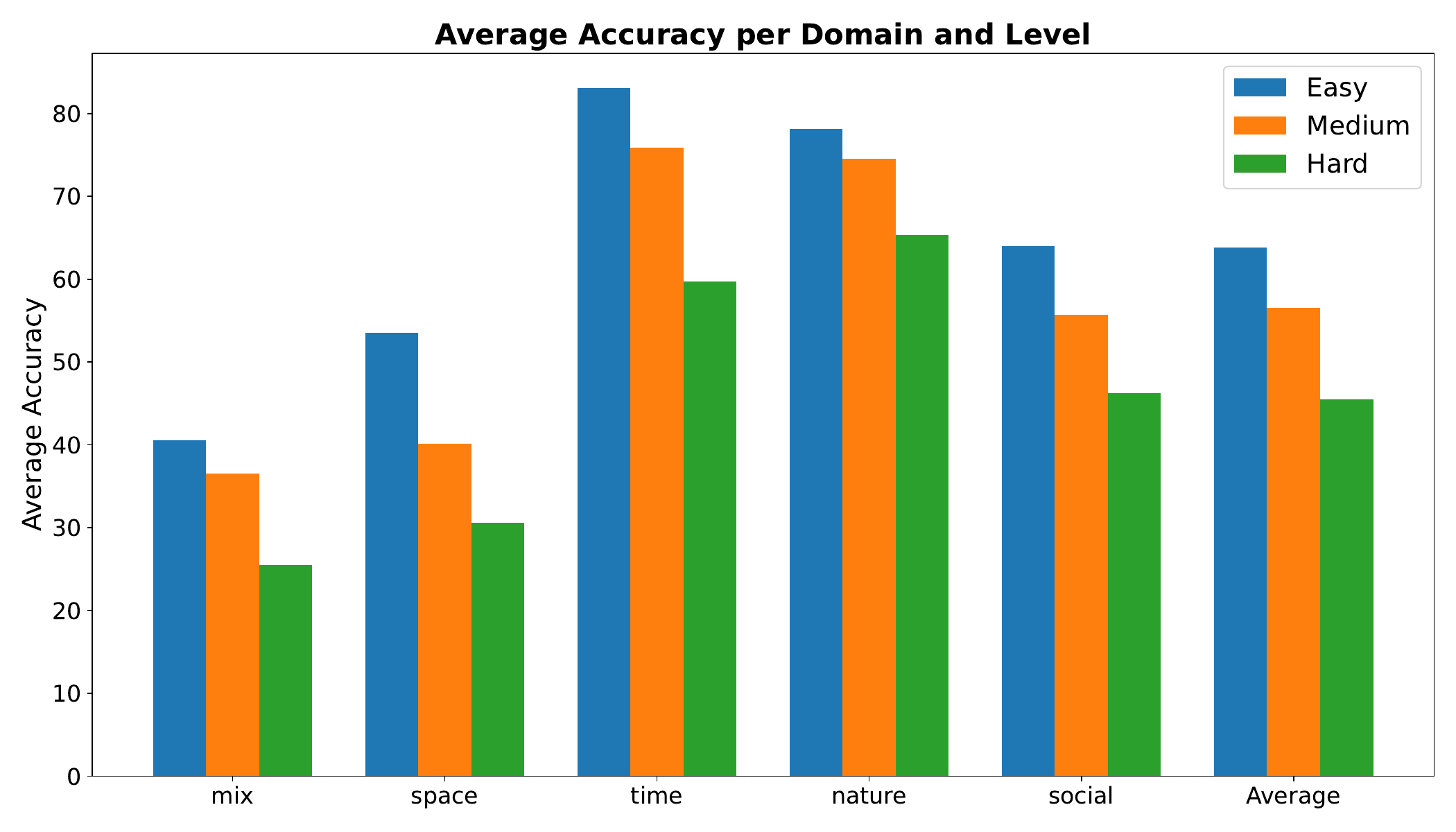}
    \caption{The average performance of LLMs on Chinese and English questions on different levels.}
    \label{fig:performance-domain-level}
\end{figure}

\paragraph{Downward Accuracy Across Difficulty Levels}
Figure \ref{fig:performance-domain-level} presents the average accuracy of different domains at three different difficulty levels: easy, medium and hard. The data shows a clear downward trend in accuracy as the difficulty increases, suggesting that the complexity of the task significantly impacts model performance.

At the easy level, the accuracy is generally high, especially in the "nature" and "time" domains, where the accuracy reaches nearly 80\%. This indicates that the models perform well on basic tasks in these areas. However, as the difficulty increases to medium level and hard level, the average accuracy of hard level is only 47.91\%. This drop reflects the increasing challenge posed by more complex reasoning tasks in these domains.
Detailed performances on different domains across difficulty levels are shown in Appendix~\ref{app:detail_performance}.

\paragraph{Cultural and Linguistic Differents Affects Performance} 
It is generally believed that large language models behave better in English than in non-English languages due to the quality and quantity of training data~\cite{zhang-etal-2023-dont}. However, on the contrary, our results show that most models perform better on Chinese questions than on English questions in space, nature and mixed domains. We filter and analyze 100 questions with the greatest performance differences between Chinese and English and find that these differences are primarily due to the following reasons:
(1) Some entities are related to geography or culture, which can result in their varying frequency across different language corpora. Although the overall size of English data is larger, the data related to these entities is more abundant in Chinese. The model performs better in understanding commonsense knowledge related to such entities in the Chinese than in English.
(2) A same word can have multiple different senses. This semantic ambiguity is more common than in English than in Chinese. The same English word may correspond to related or completely unrelated Chinese words, which may bring ambiguity to English questions. The model may confuse different senses during the reasoning process, using different senses in different steps.
Detailed examples are shown in Appendix~\ref{app:language}.

\subsection{Error Types Analysis}

\paragraph{Low-Frequency Entity Properties and Similar Social Relationships are Prone to Commonsense Errors}
The commonsense errors of LLMs can be divided into two categories: errors in entity properties and errors in entity relations. Errors in entity properties mainly occur in low-frequency entities. It is language-dependent as the frequency of an entity can vary in corpora of different languages. 
Errors in entity relations mainly manifest as the confusion of closely related, similar social relationships
 or misinterpreting asymmetric relationships.
 These types of errors reflect the limitations of LLMs in handling multifaceted commonsense knowledge. Our benchmark is designed to expose such weaknesses by featuring tasks with high knowledge point density, which require precise understanding and reasoning over multiple interconnected facts. Detailed examples are shown in Appendix~\ref{app:case_study}. 

\paragraph{Models' Internal Capabilities Limitations Lead to Reasoning Errors}
The reasoning errors of LLMs can be divided into three categories: logical contradiction, imprecision and inference error. Logical contradiction refers to the self-contradiction within different parts of the model's response.
This implies that LLMs have not yet possess logical reasoning capabilities aligned with humans. Imprecision refers to the model being affected by previous output, leading to errors in reasoning. The models may use the closer unrelated information to substitute the right information in the following analysis. The occurrence of this error is strongly related to next token prediction, which is regarded as the cornerstone of LLMs. Inference error refers to making a wrong inference based on a single clue. 
This type of error is related to the model's ability to construct and understand scenario. Our dataset is specifically designed to reveal such errors by presenting tasks with high reasoning complexity. Multi-hop reasoning instances in SCoRE effectively challenge LLMs’ logical reasoning ability. Detailed examples are provided in Appendix~\ref{app:case_study}.

\paragraph{Reasoning-Focused LLMs Tend to Overthink and Provide Additional Special Cases as Conditions}
Some reasoning LLMs, such as the OpenAI o1 and deepseek-r1 series, may overthink and invoke rare cases to fit the given conditions. For example, o1-preview argues that a birdcage can be decorated with white flowers so it can be the item with white flowers. This implies that the training methods for reasoning-focused models emphasizing logical consistency and depth of reasoning may lead them to miss more straightforward solutions. This also highlights our dataset's ability to reveal not only reasoning failures, but also cases where models over-apply complex logic and overlook intuitive solutions. Detailed examples are shown in Appendix~\ref{app:case_study}.

\paragraph{Model's Tendency in Single-Choice Questions and Logical Contradictions Lead to Answer Aggregation Errors}
Despite mentioning that our questions are multiple-choice in the prompt, the models still tend to treat them as single-choice questions. For multiple-choice questions, the models sometimes output the answer immediately after finding one correct answer. This may be due to the high frequency of single-choice questions in inference data. Additionally, logical contradictions may also contribute to errors in answer aggregation. There are examples where the model infers A is right but ultimately answers B. This demonstrates our dataset’s ability to expose errors in decision-making strategies, such as premature stopping and inconsistent answer aggregation, which are not well captured by conventional single-choice benchmarks. Detailed examples are shown in Appendix~\ref{app:case_study}.

\section{Related Work} 
According to the construction method, the previous commonsense reasoning datasets can be divided into three categories:

\paragraph{Based on Human Annotation}
These datasets are constructed through manually crafting questions and answer options, with notable examples including CommonSenseQA~\cite{talmor2019commonsenseqa}, CommonSenseQA 2.0~\cite{talmor1commonsenseqa}, ProtoQA~\cite{boratko2020protoqa}, StrategyQA ~\cite{geva2021didaristotleuselaptop} and SimpleQA~\cite{wei2024SimpleQA}. These datasets utilizes crowdsourcing approach to generate questions related to a given topic, which emphasize the quality and accuracy of data and covers a wide range of knowledge with diverse question styles. However, the high cost of manual annotation makes it difficult to build large-scale datasets using such methods.

\paragraph{Based on Template Rules}
These datasets automatically generate questions and answers through predefined task templates and rules, with notable examples including bAbI~\cite{weston2015towards}, SPARTQA~\cite{DBLP:spartqa}, StepGame~\cite{DBLP:stepgame}, SpatialEvalLLM~\cite{DBLP:spatialevalllm}, TRAM~\cite{wang2023tram}, RoomSpace~\cite{DBLP:RoomSpace} and LogicBench~\cite{parmar2024logicbench}. These datasets design templates and rules for each subtask to gather basic knowledge points.
This approach enables the rapid generation of large-scale data while significantly reducing construction costs. 
However, predefined rule templates in these datasets were limited in number and simple in structure, resulting in the lack of diversity and realism of the questions, making it difficult to ensure their generalizability. 

\paragraph{Based on LLMs}
These datasets are built by leveraging existing corpora and utilizing LLMs to generate responses or questions. mCSQA~\cite{sakai-etal-2024-mcsqa} and MuSR~\cite{sprague2024musr}.
COIG-CQIA~\cite{bai2024coig} collects questions from web data, and prompts GPT-4 to generate the responses, 
while mCSQA~\cite{sakai-etal-2024-mcsqa} and MuSR~\cite{sprague2024musr} uses LLMs to generate the whole question following pre-defined instructions. 
This approach reduces costs while enriching the diversity of questions and responses. However, the performance of current LLMs on reasoning tasks lags behind that of humans, making them unable to fully replace human input. As a result, the quality of the generated data is difficult to guarantee. Additionally, if a substantial amount of model-generated data is used for training, the performance ceiling of the trained model will be limited by the model that generated the data, making it more susceptible to model collapse during the training process.


\section{Conclusion}
In this paper, we introduce SCoRE, a bilingual dataset designed to evaluate complex logical reasoning in commonsense scenarios through multi-hop inference chains generated through a knowledge-driven synthetic data strategy. 
The dataset contains 100,000 questions across three difficulty levels and commonsense in four domains.
Our experimental results show that, despite advances in areas like code generation and math problem-solving, LLMs still face challenges in commonsense reasoning. Case studies reveal common errors, such as misinterpreting low-frequency commonsense, logical inconsistencies, and overthinking. 
The proposed data synthesis approach enables automated generation of large-scale, high-accuracy reasoning data with QA pairs, which could serve as training data to further improve LLMs' reasoning capabilities.

\section*{Limitations}

\paragraph{Scale of Knowledge Base}
Our knowledge base covers multiple domains and types of knowledge. However, the scale of the knowledge base. In the future, we plan to expand the knowledge base by adding more entities, properties, relationships, and constructing more scenarios.

\paragraph{Limited Combinations in Mix Domain}
Currently, our mix domain involves only the integration of two domains, as the knowledge base has not yet included scenarios for the integration of more domains. In the future, we plan to design more scenario types to enable diverse domain integrations.

\paragraph{Manual Error Type Analysis}
Currently, the error type analysis is conducted manually. This manual process is time-consuming and labor-intensive, and cannot comprehensively analyze all error samples. Our questions are annotated with diverse and fine-grained features, and the data synthesis process is done step by step, with a clear and traceable reasoning chain. In the future, we will explore automatic error analysis through methods such as error feature localization and reasoning path backtracking.

\section*{Ethical Consideration}
This work involved human annotation. We have provided appropriate compensation for all annotators. The total cost of annotation for the project is about 20k RMB. For all annotators, we explicitly informed them about the use of the data and required them to ensure that the questions included in SCoRE do not involve any social bias, ethical issues or privacy concerns during the annotation process.

\bibliographystyle{unsrtnat}
\bibliography{custom}

\appendix

\section{Knowledge Base Construction}
\label{app:knowledgebase}

\begin{table*}[h!] 
  \centering
  \caption{Knowledge entries written by human in four knowledge bases. The $X$ and $Y$ in the entry are placeholders for entities that can be filled in. The $A$ and $B$ are for events. The $T$ is for the times. The $V$ is for the value of natural properties.}
  \footnotesize
  \resizebox{\textwidth}{!}{
      \begin{tabular}{l | l | c| p{10cm}}
        \toprule
        \textbf{Domain} & \textbf{Type of Entry} & \textbf{\#} & \textbf{Example of Entry}\\
        \midrule
        
        \multirow{6}{*}{Space} & Spatial Property & 59 & $X$ faces south, $X$ is on the first floor, etc.\\
        \cline{2-4}
        & Spatial Relation & 244 & $X$ is to the left of $Y$, $X$ is facing away from $Y$, etc. \\
        \cline{2-4}
        & Logical Rule & 861 & \makecell*[l]{"$X$ is to the left of $Y$" is equivalent to "$Y$ is to the right of $X$";\\"$X$ is on the first tier" and "$Y$ is on the third tier" can imply that "$X$ is separated \\from $Y$ by one tier in between".}\\
        \midrule
        
        \multirow{7}{*}{Time} & Temporal Event & 39 & $X$ get married in $T$, $X$ played badminton on $T$, etc. \\
        \cline{2-4}
        & Temporal Relation & 8 & $A$ happened earlier than $B$, $A$ happened $T$ days after $B$, etc.\\
        \cline{2-4}
        & Logical Rule & 51 & \makecell*[l]{"$A$ happened before $B$" is equivalent to "$A$ happended after $B$"; \\"$A$ happened in $T$" and "$B$ happened in $T$" can imply that "$A$ and $B$ happened \\at the same time"; \\In a person's life, getting married happens later than starting elementary school.}\\
        
        \midrule
        \multirow{2}{*}{Social} & Individual Property & 42 & surname:Li, first name:Dawei, gender:male \\
        \cline{2-4}
        & Social Relation & 18 & $X$ is $Y$'s father, $X$ is a friend of $Y$, etc. \\
        
        \midrule
        \multirow{5}{*}{Nature} & Natural Property & 16 & $X$ is a kind of $V$, $X$ has $V$ legs, etc.\\
        \cline{2-4}
        & Comparative Relation & 10 & $X$ has a longer wavelength than the light reflected by $Y$.\\ \cline{2-4}
        & Logical Rule & 27 & \makecell*[l]{"$X$ is a bird" can imply that "$X$ is a homothermal animal",\\red light has a longer wavelength than yellow light,\\A same stone sinks faster in freshwater than in seawater.\\}\\
        \bottomrule
      \end{tabular}
    }
  \label{tab:human_number}
\end{table*}

The knowledge bases in four domains are constructed under the same overarching knowledge framework, which includes entities, scenarios, and propositions to express properties and relations. We first extracted knowledge from existing external resources, then manually created proposition templates. These templates, once filled up, will generate factual statements during the data synthesis process. The logical rules among the templates needed by the Reasoner are also connected manually. After three rounds of manual check, we formed our current knowledge bases.

\paragraph{External Resources}
In time domain, 9 historical events that occurred between 1900 and 2000, such as the birth year of Michael Jackson, were selected from Wikipedia and added to the knowledge base. In social domain, 58 complex family relationships spanning two to three generations are comes from a calculator\footnote{https://github.com/mumuy/relationship/} for computing Chinese social relationships, such as grandfather is equivalent to father's father. In nature domain, 633 entities are from HowNet and then get their properties from HowNet~\cite{dong2003hownet}, ConceptNet\footnote{These resources are used under the CC-BY-4.0 license.}~\cite{speer2017conceptnet} and the Contemporary Chinese Dictionary. The properties are clustered according to the Attribute Value list in HowNet.

\paragraph{Manual Construction}
Table \ref{tab:human_number} presents the number of knowledge entries written by humans in four knowledge bases. Scenarios are shown in Appendix \ref{app:scenario_diagram}.

\begin{table}[h] 
  \centering
  \caption{Examples of the property of statements in the knowledge base}
  \footnotesize
      \begin{tabular}{p{1.2cm} | p{2.3cm} | p{1.8cm} | l}
        \toprule
        \textbf{Property} & \textbf{Definition} & \textbf{Statement} & \textbf{Value}\\
        \midrule
        truth value & Truth value of a statement, either true or false & Tom isn't the son of Thomas. & false\\
        \midrule
        value precision & Number of values determined by the statement, either precise (single-valued) or vague (multi-valued). & Jack is within the range to the left of Tom. & vague \\
        \midrule
        time & Time that a event in the statement happens & Mary retired in 1955. & 1955 \\
        \bottomrule
      \end{tabular}
  \label{tab:example_relation}
\end{table}

\section{Properties of statements}
\label{app:examples_relation_property}
Our knowledge bases define certain properties to refine the context of statements, as listed in table \ref{tab:example_relation}.

\section{Human Verification and Refinement}
\label{app:human_verify}
Human oversight is a quality control measure for the knowledge base. We conducted a total of three rounds of human verification.

In the first round, we directly audited the knowledge base. Eight graduate students were recruited to review all knowledge entries. Approximately \textbf{15\%} of the entries were revised based on their evaluation.

In the second round, we performed a stratified sampling inspection on the synthesized data. For each domain, 10\% of the generated data was randomly sampled. The sampled items spanned all difficulty levels, scenarios, and question types, covering as many distinct knowledge entries as possible. Ten graduate students participated in this round of verification. They were instructed to assess the following aspects:

\renewcommand{\labelitemi}{-}
\begin{itemize}
    \item Correctness of the question: whether there were contradictions, commonsense violations, or missing conditions;
    \item Correctness of the answer: whether the answer of the questions are correct;
    \item Linguistic Quality: whether the questions were expressed clearly and naturally, rated on a 3-point scale.
\end{itemize}

The results showed that approximately \textbf{30\%} of the questions had issues in at least one of the above three dimensions. We traced each error back to its originating knowledge entry and made the necessary corrections.

The third round followed the same procedure as the second, focusing on evaluating the correctness of the question and answers. This round revealed that, with the exception of the social domain, where errors were caused by the reasoning program, all other domains achieved \textbf{100\%} correctness. After fixing the errors of the program, both the dataset and the knowledge base reached complete accuracy.

\section{Question examples of different domains and question-types}
\label{app:question_examples}
Table~\ref{tab:question_example} shows the examples of questions across different domains and question-types in our dataset.

\footnotesize

\begin{longtable}{l | p{1.5cm} | p{1.5cm} |p{8cm}}
\toprule \textbf{Domain} & \textbf{Scenario} &\textbf{Question type} & \textbf{Example of question} \\ \midrule
\endfirsthead
\toprule \textbf{Domain} & \textbf{Scenario} &\textbf{Question type} & \textbf{Example of question} \\ \midrule
\endhead
\bottomrule
\endfoot
\endlastfoot

\textbf{Space}
& \multirow{2}{=}{centrifugal hexagon} 
& 6 slots - vague & 

周伯通、郝大通、柯镇恶、赵志敬、刘处玄、王重阳六位道士在终南山重阳宫内盘腿席地打坐，围成一个圆圈，修炼内功，六人的位置恰好形成一个正六边形。六人都面朝外背对圆心而坐。任意相邻两人之间的间距相等，大约为一米。已知：\\
&&& 刘处玄的右边接着就是赵志敬，\\
&&& 郝大通在赵志敬的右边，二者相邻，\\
&&& 从赵志敬的左边数起第二个位置是王重阳，\\
&&& 从柯镇恶的左边数起第五个位置是王重阳，\\
&&& 从王重阳的左边数起第二个位置是周伯通。\\
&&& 问题：\\
&&& 赵志敬与\_\_\_之间隔着两个位置。\\
&&& 选项：A.刘处玄 \quad B.周伯通 \quad C.柯镇恶 \quad D.郝大通\\
&&& 答案：C\\
&&& David, Jennifer, Elizabeth, Michael, John, James, —these six Taoist priests are seated cross-legged on the ground inside the Chongyang Palace on Zhongnan Mountain, arranged in a circle as they practice internal martial arts. The positions of the six priests form a perfect hexagon. Each priest is facing outward, with their backs toward the center of the circle. The distance between any two adjacent priests is equal, approximately one meter. It is known that: \\
&&& Michael is directly to the right of John; \\
&&& Jennifer is directly to the right of Michael; \\
&&& James occupies the second position to the left of Michael; \\
&&& James occupies the fifth position to the left of Elizabeth;\\
&&& David occupies the second position to the left of James. \\
&&& question:\\
&&& Michael and \_\_\_ are separated by two positions.\\
&&& Options: A.John \quad B.David \quad C.Elizabeth \quad D.Jennifer \\
&&& Answer:C
\\ \cline{3-4}
& & 5 slots - precise & 

柯镇恶、王处一、郝大通、尹志平、刘处玄、赵志敬六位道士在终南山重阳宫内盘腿席地打坐，围成一个圆圈，修炼内功，六人的位置恰好形成一个正六边形。六人都面朝外背对圆心而坐。任意相邻两人之间的间距相等，大约为一米。已知：\\
&&& 尹志平在赵志敬右边数起第五个位置，\\ 
&&& 柯镇恶的左边紧接着就是刘处玄，\\
&&& 尹志平在郝大通左边数起第二个位置。\\
&&& 问题：\\
&&& \_\_\_在尹志平左边数起第三个位置。\\
&&& 答案：D\\
&&& 选项：A.郝大通 \quad B.刘处玄 \quad C.柯镇恶 \quad D.以上选项都不是\\
&&& Elizabeth, Robert, Jennifer, Susan, John, Michael, —these six Taoist priests are seated cross-legged on the ground inside the Chongyang Palace on Zhongnan Mountain, arranged in a circle as they practice internal martial arts. The positions of the six priests form a perfect hexagon. Each priest is facing outward, with their backs toward the center of the circle. The distance between any two adjacent priests is equal, approximately one meter. It is known that: \\
&&& Susan occupies the fifth position to the right of Michael; \\
&&& John is directly to the left of Elizabeth; \\
&&& Susan occupies the second position to the left of Jennifer.\\   
&&& question:\\
&&& \_\_\_ occupies the third position to the left of Susan.\\
&&& Options: A.Jennifer \quad B.John \quad C.Elizabeth \quad D.None of the above \\
&&& Answer:D 
\\ \cline{2-4}
& \multirow{2}{=}{three rows two columns}
& 6 slots - vague & 

月季、水仙、茉莉、君子兰、天竺葵、郁金香六盆花放置在三层花架上呈列，花架紧靠大厅南墙放置，每层两格，各放一盆花，一在东，一在西。画师站在花架前，面对花架支起画架，为花架中六盆花画素描。在描述各花的方位关系时，约定以画师自身左右方位为参照，即东侧花盆为左，西侧花盆为右。东侧花盘在西侧花盘左边，西侧花盆在东侧花盆右边。已知：\\
&&& 郁金香的正上方是月季的正下方，\\ 
&&& 月季在天竺葵左上方且二者隔了一层，\\
&&& 天竺葵在一层西侧，\\
&&& 君子兰在天竺葵左上方且二者不隔层，\\
&&& 月季在茉莉左边，\\
&&& 水仙在君子兰右边。\\
&&& 问题：\\
&&& \_\_\_所在层和天竺葵所在层相邻。\\
&&& 答案：D\\
&&& 选项：A.茉莉 \quad B.月季 \quad C.郁金香 \quad D.以上选项都不是\\
&&& Monthly Rose, Narcissus, Jasmine, Clivia, Geranium, Tulip, —these six pots of flowers are arranged on a three-tiered flower stand, placed against the south wall of the hall. Each tier is divided into two sections, with one pot placed in each section—one on the east side and one on the west. The artist stands in front of the flower stand, facing it, and sets up an easel to sketch the six pots of flowers. When describing the positional relationships of the flowers, the artist's own left and right are used as a reference, with the eastern pot being on the left and the western pot being on the right. In other words, the eastern pot is to the left of the western pot, and the western pot is to the right of the eastern pot. It is known that:\\
&&& The position directly above Tulip is directly below Monthly Rose; \\
&&&Monthly Rose is located at the upper left side of Geranium and there is a tier between them;\\
&&& Geranium is on the west side of the first floor; \\
&&& Clivia is located in the upper left corner of Geranium and the two are not separated by a layer; \\
&&& Monthly Rose is to the left of Jasmine;\\
&&& Narcissus is somewhere to the right of Clivia. \\
&&& question:\\
&&& The tier where \_\_\_ is located above or below the tier where Geranium is located.\\
&&& Options: A.Jasmine \quad B.Monthly Rose \quad C.Tulip \quad D.None of the above \\
&&& Answer:D

\\ \cline{3-4}
& & 5 slots - precise & 

波斯菊、天竺葵、君子兰、郁金香、茉莉、茶花六盆花放置在三层花架上呈列，花架紧靠大厅南墙放置，每层两格，各放一盆花，一在东，一在西。画师站在花架前，面对花架支起画架，为花架中六盆花画素描。在描述各花的方位关系时，约定以画师自身左右方位为参照，即东侧花盆为左，西侧花盆为右。东侧花盘在西侧花盘左边，西侧花盆在东侧花盆右边。已知：\\
&&& 茉莉在二层左侧，\\ 
&&& 郁金香在茶花左边且二者同层，\\
&&& 茉莉在郁金香正上方且二者不隔层，\\
&&& 茉莉在波斯菊左下方且二者不隔层，\\
&&& 君子兰的右邻在茉莉的右上方。\\
&&& 问题：\\
&&& \_\_\_在茉莉右边且二者同层。\\
&&& 答案：D\\
&&& 选项：A.郁金香 \quad B.君子兰 \quad C.波斯菊 \quad D.以上选项都不是\\
&&& Cosmos, Geranium, Clivia, Tulip, Jasmine, Camellia, —these six pots of flowers are arranged on a three-tiered flower stand, placed against the south wall of the hall. Each tier is divided into two sections, with one pot placed in each section—one on the east side and one on the west. The artist stands in front of the flower stand, facing it, and sets up an easel to sketch the six pots of flowers. When describing the positional relationships of the flowers, the artist's own left and right are used as a reference, with the eastern pot being on the left and the western pot being on the right. In other words, the eastern pot is to the left of the western pot, and the western pot is to the right of the eastern pot. It is known that:\\
&&& Jasmine is on the left side of the second tier;\\
&&& Tulip is to the left of Camellia and both are on the same level;\\
&&& Jasmine is directly above Tulip and they are not seperated by any tier;\\
&&& Jasmine is located at the lower left of Cosmos and there is no tier between them; \\
&&& The right side neighbor of Clivia is above the upper right side of Jasmine. \\

&&& question:\\
&&& \_\_\_ is to the right of Jasmine and both are on the same tier.\\
&&& Options: A.Tulip \quad B.Clivia \quad C.Cosmos \quad D.None of the above \\
&&& Answer:D 

\\ 

\midrule

\textbf{Time} & \multirow{1}{=}{Linear Scenario} &  \multirow{1}{=}{Precise Event} & 小明的女儿正在给朋友讲述父亲的一生: \\
&&& (1)在迈克尔·杰克逊出生之前58年，他出生; \\
&&& (2)他度过一生的时长为99年; \\
&&& (3)在联合国成立的38年之前，他开始上小学; \\
&&& (4)他小学毕业的时间比迈克尔·杰克逊出生早45年; \\
&&& (5)在1913年，他开始上初中; \\
&&& (6)他上初中一共3年; \\
&&& (7)在联合国成立的30年之前，他遇见未来的妻子; \\
&&& (8)他开始上高中的时间比他出生晚16年; \\
&&& (9)他高中毕业的时间比他开始上初中晚6年; \\
&&& (10)在他遇见未来的妻子之后4年，他开始上大学; \\
&&& (11)在他初中毕业之后8年，他大学毕业。\\ 
&&& 问题：\_\_\_\_是在迈克尔·杰克逊出生的42年前。\\
&&& 选项：A.他开始上小学 \quad B.他去世 \quad C.他初中毕业 \quad D.他高中毕业 \\
&&& 答案：C \\
&&& Jack's daughter is telling her friends about the story of her father's life: \\
&&& (1)Jack was born 58 years before Michael Jackson was born; \\
&&& (2)Jack lived his life for 99 years; \\
&&& (3)Jack started elementary school 38 years before the United Nations was founded; \\
&&& (4)Jack graduated from elementary school 45 years before Michael Jackson was born; \\
&&& (5)Jack started junior high school in 1913; \\
&&& (6)Jack studied in junior high school for 3 years; \\
&&& (7)Jack met his future wife 30 years before the United Nations was founded; \\
&&& (8)Jack started high school 16 years after he was born; \\
&&& (9)Jack graduated from high school 6 years after he started junior high school; \\
&&& (10)Jack started university 4 years after he met his future wife; \\
&&& (11)Jack graduated from university 8 years after he graduated from junior high school. \\
&&& Question: \_\_\_\_ 42 years before Michael Jackson was born. \\
&&& Options: \\
&&& A.Jack started elementary school \\
&&& B.Jack passed away \\
&&& C.Jack graduated from junior high school \\
&&& D.Jack graduated from high school \\
&&& Answer: C \\
\cline{3-4}
&  &  \multirow{1}{=}{Correct Statements} & 小明的女儿正在给朋友讲述父亲的一生: \\
&&& (1)在1900年，他出生; \\
&&& (2)在联合国成立的54年之后，他去世; \\
&&& (3)他开始上初中的时间比第一届普利策奖颁发早4年; \\
&&& (4)在迈克尔·杰克逊出生的42年之前，他初中毕业; \\
&&& (5)在1915年，他遇见未来的妻子; \\
&&& (6)他开始上高中的时间比他出生晚16年; \\
&&& (7)在他开始上初中之后6年，他高中毕业; \\
&&& (8)他成为父亲的时间比他遇见未来的妻子晚23年; \\
&&& (9)他退休的时间比他出生晚64年。\\
&&& 问题：以下选项中正确的是\_\_\_\_ \\
&&& 选项：\\
&&& A. 谷歌创立和联合国成立之间相隔97年。\\
&&& B. 他成为父亲的时间比谷歌创立早20年。\\
&&& C. 他开始上初中是在第一届普利策奖颁发之后。\\
&&& D. 他开始上高中和他出生之间相隔16年。\\
&&& 答案：D\\
&&& Jack's daughter is telling her friends about the story of her father's life: \\
&&& (1)Jack was born in 1900; \\
&&& (2)Jack passed away 54 years after the United Nations was founded; \\
&&& (3)Jack started junior high school 4 years before the first Pulitzer Prizes were announced; \\
&&& (4)Jack graduated from junior high school 42 years before Michael Jackson was born; \\
&&& (5)Jack met his future wife in 1915; \\
&&& (6)Jack started high school 16 years after he was born; \\
&&&(7)Jack graduated from high school 6 years after he started junior high school; \\
&&& (8)Jack became a father 23 years after he met his future wife; \\
&&& (9)Jack retired 64 years after he was born. \\
&&& Question: Select the correct statement(s): \_\_\_\_\\ 
&&& Options: \\
&&& A.The gap between the time the United Nations was founded and the time Google was founded is 97 years. \\
&&& B.Jack became a father 20 years before Google was founded. \\
&&& C.Jack started junior high school after the first Pulitzer Prizes were announced. \\
&&& D.The gap between the time Jack was born and the time he started high school is 16 years. \\
&&& Answer: D \\ \cline{2-4}
& \multirow{1}{=}{Cyclic Scenario} & \multirow{1}{=}{Incorrect Statements} & 小明是一名大学生，以下是他的每周安排: \\
&&&(1)星期三，他打羽毛球; \\
&&&(2)周三，他开组会; \\
&&&(3)在星期三，他跑步; \\
&&&(4)在他打羽毛球之后1天，他阅读科幻小说; \\
&&&(5)在他开组会的2天之后，他练习吉他; \\
&&&(6)在他开组会之后3天，他看论文。\\
&&& 问题：以下选项中不正确的是\_\_\_\_ \\
&&& 选项： \\
&&& A.在他打羽毛球的5天之前，他跑步。\\
&&& B.在他阅读科幻小说的6天之后，他打羽毛球。\\
&&& C.在他看论文的3天之前，他阅读科幻小说。\\
&&& D.在他练习吉他之后2天，他阅读科幻小说。\\
&&& 答案：ACD \\
&&& Jack is a college student, and here are his weekly plans: \\
&&&(1)On Wednesday, Jack plays badminton; \\
&&&(2)Jack has a group meeting on Wednesday; \\
&&&(3)Jack goes jogging on Wednesday; \\
&&&(4)Jack reads sci-fi novels 1 day after he plays badminton; \\
&&&(5)2 days after Jack has a group meeting, he practices the guitar; \\
&&&(6)3 days after Jack has a group meeting, he reads research papers. \\
&&& Question: Select the incorrect statement(s): \_\_\_\_ \\
&&& Options: \\
&&& A.Jack goes jogging 5 days before he plays badminton. \\
&&& B.Jack plays badminton 6 days after he reads sci-fi novels. \\
&&& C.Jack reads sci-fi novels 3 days before he reads research papers. \\
&&& D.2 days after Jack practices the guitar, he reads sci-fi novels. \\
&&& Answer: ACD \\
\midrule

\textbf{Social} &  \multirow{1}{=}{General Social Relationship} &  \multirow{1}{=}{Correct statements} & 已知：冯志强是冯秀英的哥哥，也是周强的朋友。王丽是冯志强的妻子，也是周强的同学。郑建国是王丽的前男友，也是李晓静的男朋友。吴晶是郑建国的数位前女友中的一位，也是孙大伟的前妻。周强是吴晶的领导，也是赵伟的下属。李晓静是周强的数位领导中的一位，也是孙大伟的同事。钱静是孙大伟的前女友，也是赵伟的女朋友。\\
&&&问题：以下选项正确的是\_\_\_ \\ 
&&&选项：\\
&&&A.孙大伟的同事是吴晶的前男友的媳妇 \\
&&&B.孙大伟的前妻是李晓静的男朋友的母亲\\
&&&C.赵伟的下属是郑建国的某位名字中带有丽的前女友的同学\\
&&&D.赵伟的下属是郑建国的某位名字中带有丽的前女友的同事\\
&&&答案：C\\
&&&(Known: Feng Zhiqiang is Feng Xiuying's elder brother and also a friend of Zhou Qiang. Wang Li is the wife of Feng Zhiqiang and also a classmate of Zhou Qiang. Zheng Jianguo is Wang Li's ex boyfriend and also Li Xiaojing's boyfriend. Wu Jing is one of Zheng Jianguo's several ex girlfriends and also the ex-wife of Sun Dawei. Zhou Qiang is the leader of Wu Jing and also a subordinate of Zhao Wei. Li Xiaojing is one of Zhou Qiang's digital leaders and also a colleague of Sun Dawei. Qian Jing is the ex girlfriend of Sun Dawei and also the girlfriend of Zhao Wei\\
&&&Question: Which of the following options is correct\\
&&&Options:\\
&&&A. Sun Dawei's colleague is the wife of Wu Jing's ex boyfriend\\
&&&B. Sun Dawei's ex-wife is the mother of Li Xiaojing's boyfriend\\
&&&C. Zhao Wei's subordinate is a classmate of Zheng Jianguo whose name includes Li's ex girlfriend\\
&&&D. Zhao Wei's subordinate is a colleague of Zheng Jianguo whose name includes Li's ex girlfriend\\
&&&Answer: C)\\
\cline{2-4}
& Chinese Family Relationship & Incorrect statement & 已知：赵丽是赵静的外孙女，也是赵强的姐姐的外孙女。赵力是赵军的妈妈的爷爷，也是赵晓丽的外孙。赵秀英是赵强的女儿，也是赵晓丽的姐姐。赵秀兰是赵强的妹妹。赵晓静是赵静的女儿。赵石是赵军的外公，也是赵玉兰的姐姐的侄女的外甥女的孙子。赵丽是赵晓丽的姐姐的大姑的外孙女。赵琳是赵军的孙女。赵玉兰是赵秀英的爸爸的妹妹。赵芳是赵鹏的奶奶，也是赵盈盈的孙子的女儿。赵石是赵鹏的爸爸的外公。\\
&&&问题：以下选项不正确的是\_\_\_ \\ 
&&&选项：\\
&&&A.赵军是赵琳的爷爷。\\
&&&B.赵晓丽是赵鹏的奶奶的爷爷的外婆。\\
&&&C.赵秀兰是赵晓丽的侄女的妹妹。\\
&&&D.赵军是赵秀英的外甥女的孙子的外孙。\\
&&&答案：C\\
&&&(Known:Zhao Li is the granddaughter(daughter's daughter) of Zhao Jing and also the granddaughter(daughter's daughter) of Zhao Qiang's elder sister. Zhao Li is the grandfather(father's father) of Zhao Jun's mother and also the grandson(son's son) of Zhao Xiaoli. Zhao Xiuying is the daughter of Zhao Qiang and also the elder sister of Zhao Xiaoli. Zhao Xiulan is Zhao Qiang's younger sister. Zhao Xiaojing is Zhao Jing's daughter. Zhao Shi is the grandfather(mother's father) of Zhao Jun and the grandson(son's son) of the niece(sister's daughter) of Zhao Yulan's elder sister's niece(brother's daughter). Zhao Li is the granddaughter(daughter's daughter) of Zhao Xiaoli's elder sister's aunt(father's sister). Zhao Lin is the granddaughter(son's daughter) of Zhao Jun. Zhao Yulan is the younger sister of Zhao Xiuying's father. Zhao Fang is Zhao Peng's grandmother(father's mother) and also the daughter of Zhao Yingying's grandson(son's son). Zhao Shi is the grandfather(mother's father) of Zhao Peng's father.\\
&&&Question: Which of the following options is incorrect\\
&&&Options:\\
&&&A. Zhao Jun is Zhao Lin's grandfather(father's father).\\
&&&B. Zhao Xiaoli is the grandmother(mother's mother) of Zhao Peng's grandmother(father's mother)'s grandfather(father's father).\\
&&&C. Zhao Xiulan is the younger sister of Zhao Xiaoli's niece(brother's daughter).\\
&&&D. Zhao Jun is the grandson(daughter's son) of Zhao Xiuying's niece(sister's daughter)'s grandson(son's son)'s grandson(daughter's son).\\
&&&Answer: C)\\
\midrule

\textbf{Nature} & \multirow{1}{=}{Farming} & \multirow{1}{=}{Precise entity}& 一位勤劳的农夫有四块田，他在四块田中分别种植了南瓜、开心果、瓠子、枇杷四种作物。已知：\\
&&& 1号田中的作物是一种坚果；\\
&&& 2号田中的作物属于蔬菜；\\
&&& 3号田中的作物反射的光比2号田中的作物反射的光波长更长；\\
&&& 3号田中的作物属于蔬菜。\\
&&& 问题：4号田中种的是\_\_\_\_。\\
&&& 选项：A.南瓜 B.开心果 C.瓠子 D.枇杷 \\
&&& 答案：D \\
&&& A hardworking farmer has four fields. In each field he plants one of four crops: pumpkin, pistachio nut, edible gourd, loquat. It is known that: The plant in field No.1 is a kind of nut; The plant in field No.2 is a vegetable; The light reflected by the plant in field No.3 has a longer wavelength than the light reflected by the plant in field No.2; The plant in field No.3 is a vegetable.\\
&&& Question: \_\_\_\_ is planted in field No.4.\\
&&& Options: A.pumpkin B.pistachio nut C.edible gourd D.loquat\\
&&& Answer: D\\
 \cline{2-4}
& Zoo enclosures & Precise position & 动物园里的四个场馆分别养着蝴蝶、章鱼、水牛、白鹭四种动物。已知：\\
&&&3号场馆中的动物比2号场馆中的动物多2条腿；\\
&&&3号场馆中的动物属于恒温动物；\\
&&&4号场馆中的动物比3号场馆中的动物多2条腿。\\
&&&问题：白鹭养在\_\_\_\_号场馆中。\\
&&&选项: A.1 B.2 C.3 D.4\\
&&&答案：B\\ 
&&&The four enclosures in the zoo keep four different kinds of animals: butterfly, octopus, buffalo, egret. Now we know that: The animal in enclosure No.3 has 2 more legs than the animal in enclosure No.2; The animal in enclosure No.3 is a homothermal animal; The animal in enclosure No.4 has 2 more legs than the animal in enclosure No.3.\\

\cline{2-4}
& \multirow{1}{=}{Items on photos} & \multirow{1}{=}{vague position} & 一面墙上贴着芒果、苹果汁、杨梅、胡萝卜四种物品的照片。已知：\\
&&&4号照片上中的物品不属于水果；\\
&&&3号照片上的物品反射的光比1号照片上的物品反射的光波长更长；\\
&&&2号照片上的物品的味道是酸的；\\
&&&4号照片上中的物品不属于蔬菜；\\
&&&2号照片上的物品属于水果。\\
&&&问题：水果在\_\_\_\_号照片上。\\
&&&选项: A.1 B.2 C.3 D.4\\
&&&答案: AB\\
&&&On a wall pasted photos of four different items: mango, apple juice, bayberry, carrot. Now we know that: \\
&&&The item on photo No.4 is not a fruit; \\
&&&The light reflected by the item on photo No.3 has a longer wavelength than the light reflected by the item on photo No.1; \\
&&&The item on photo No.2 tastes sour; \\
&&&The item on photo No.4 is not a vegetable; \\
&&&The item on photo No.2 is a fruit. \\
&&&Question: A fruit is on photo No.\_\_\_\_.\\
&&&Options: A.1 B.2 C.3 D.4\\
&&&Answer: AB\\
\midrule
\textbf{Mix} & Three rows two columns & 5 slots - precise& 香草、铅笔盒、吐司、缝衣针、大葱、花生糖六种商品在三层货架上放置，货架紧靠商店南墙放置，每层两格，各放一种商品，一在东，一在西。顾客站在货架前选购商品。在描述各商品的位置关系时，约定以顾客自身左右方位为参照，即东侧商品为左，西侧商品为右。已知：\\
&&&属于调味料的蔬菜在花草正下方且二者隔了一层；\\
&&&花草在顶层；\\
&&&花草在吐司左边；\\
&&&甜的加工食品在吐司左下方且二者不隔层；\\
&&&铅笔盒和属于调味料的蔬菜在同一层。\\
&&& 问题：\_\_\_在铅笔盒左上方且二者不隔层 \\
&&& 选项：A.吐司 B.香草 C.大葱 D.以上选项都不是\\
&&&答案：D\\
&&& Vanilla, pencil-box, toast, sewing needle, scallion, peanut brittle, six items are placed on a three-tier shelf, which is positioned against the south wall of the store. Each tier has two sections, with one type of item placed in the east section and one in the west section. A customer is standing in front of the shelf. When describing the positional relationships of the items, it is agreed that the customer's own left and right will be used as a reference, with the east section being on the left and the west section being on the right. It is known that:\\
&&&The seasoner and vegetable is located directly below the flower or grass and separated by one tier,\\
&&&The flower or grass is on the top floor,\\
&&&The flower or grass is to the left of the toast,\\
&&&The sweet processed food is located at the lower left of the toast and there is no tier between them,\\
&&&The pencil-box and the seasoner and vegetable are on the same level.\\
&&&\_\_\_ is located in the upper left corner of pencil-box and the two are not separated by a layer\\
&&&Options: A.toast B.vanilla C.scallion D. None of the above\\
&&&Answer: D\\
\cline{2-4}
& Three rows two columns & 6 slots - vague &南瓜、白鲢、红豆、柠檬、购物袋、龙须面六种商品在三层货架上放置，货架紧靠商店南墙放置，每层两格，各放一种商品，一在东，一在西。顾客站在货架前选购商品。在描述各商品的位置关系时，约定以顾客自身左右方位为参照，即东侧商品为左，西侧商品为右。已知：\\
&&&加工食品在鱼上一层；\\
&&&工具在顶层；\\
&&&一层左侧是开黄色花的橙色的物品；\\
&&&加工食品在黄色的物品正上方；\\
&&&红色的物品在黄色的物品正下方；\\
&&&二层西侧是黄色的物品。\\
&&& 问题：\_\_\_红豆上方 \\
&&& 选项：A.白鲢 B.龙须面 C.柠檬 D.以上选项都不是\\
&&& 答案：ABC\\
&&&Pumpkin, silver carp, red bean, lemon, carrier bag, dragon whiskers noodles, six items are placed on a three-tier shelf, which is positioned against the south wall of the store. Each tier has two sections, with one type of item placed in the east section and one in the west section. A customer is standing in front of the shelf. When describing the positional relationships of the items, it is agreed that the customer's own left and right will be used as a reference, with the east section being on the left and the west section being on the right. It is known that:\\
&&&The processed food is one tier above the fish,\\
&&&The tool is on the top floor,\\
&&&The orange item with yellow flower is on the left side of the first tier,\\
&&&The processed food is above the yellow item in a straight line,\\
&&&The red item is directly under the yellow item,\\
&&&The yellow item is on the west side of the second floor.\\
&&&\_\_\_ is above the red bean\\
&&&Options: A.silver carp B.dragon whiskers noodles C.lemon D. None of the above\\
&&&Answer: ABC\\
\bottomrule
\label{tab:question_example}
\end{longtable}

\section{Difficulty Calculation Formula of Reasoning Problems in Various Domains} 
\label{app:level}

All domain-specific difficulty formulas follow a unified design philosophy: they combine weighted indicators reflecting cognitive load (e.g., entity count, reasoning chain length), knowledge complexity (e.g., property difficulty, scene type), and information completeness (e.g., explicit mentions, coverage). The coefficients balance domain-specific priorities while maintaining comparable difficulty scales across categories.

\subsection*{Domain-Specific Formulas:}

    \noindent \textbf{Space:}
    \begin{align*}
        level = 0.3nm + 0.5pr + 0.2al
    \end{align*}
    \renewcommand{\labelitemi}{-}
    \begin{itemize}
        \item $nm$: Number of entities
        \item $pr$: Precision, $pr=1$ when the number of entities involved in the question exceeds 2
        \item $al$: Whether entity coverage of the text is incomplete, $al=1$ when entity coverage is incomplete
    \end{itemize}

    \noindent \textbf{Time:}
    \begin{align*}
        level = 0.02lc + 0.2cg + 0.25na 
        + 0.05dk + 0.5dq + dc
    \end{align*}
    \begin{itemize}
        \item $lc$: The length of COT
        \item $cg$: Difficulty of the most challenging statement
        \item $na$: Number of options
        \item $dk$: Difficulty of knowledge
        \item $dc$: Difficulty of scenario, $dc=0.5$ for cyclic scenarios, 0 otherwise
        \item $dq$: Difficulty of core question
    \end{itemize}

   \noindent \textbf{Social:} 
    \begin{align*}
        level = 0.4cl + 0.3nm + 0.3ne
    \end{align*}
    \begin{itemize}
        \item $cl$: Reasoning chain length
        \item $nm$: Whether the entities in the question are explicitly mentioned in the text, $nm=1$ when entities are not explicitly mentioned
        \item $ne$: Relationship edges in network
    \end{itemize}

    \noindent \textbf{Nature:} 
    \begin{align*}
        level = 0.4kl + 0.3cl + 0.5nm
    \end{align*}
    \begin{itemize}
        \item $kl$: The sum of property difficulty
        \item $cl$: Reasoning chain length
        \item $nm$: Whether the entity in the question is mentioned in the text, $nm=1$ when entities are not mentioned
    \end{itemize}

    \noindent \textbf{Mix:} 
    \begin{align*}
        level = 0.4level_{\text{nature}} + level_{\text{space}} + \frac{nm_q + nm_a}{2}
    \end{align*}
    \begin{itemize}
        \item $level_{\text{space}}$: Space difficulty, $level_{\text{space}}=1$ for vague spatial templates
        \item $level_{\text{nature}}$: Nature difficulty, equal to $kl$ in nature domain
        \item $nm_q/nm_a$: Whether all entities in the question/answer are explicitly mentioned in the text, $nm_q/nm_a=1$ when at least of entity in the question/answer are implicit
    \end{itemize}

\section{Experiments Setting}
\label{app:experiments setting}
\begin{figure}[h!]
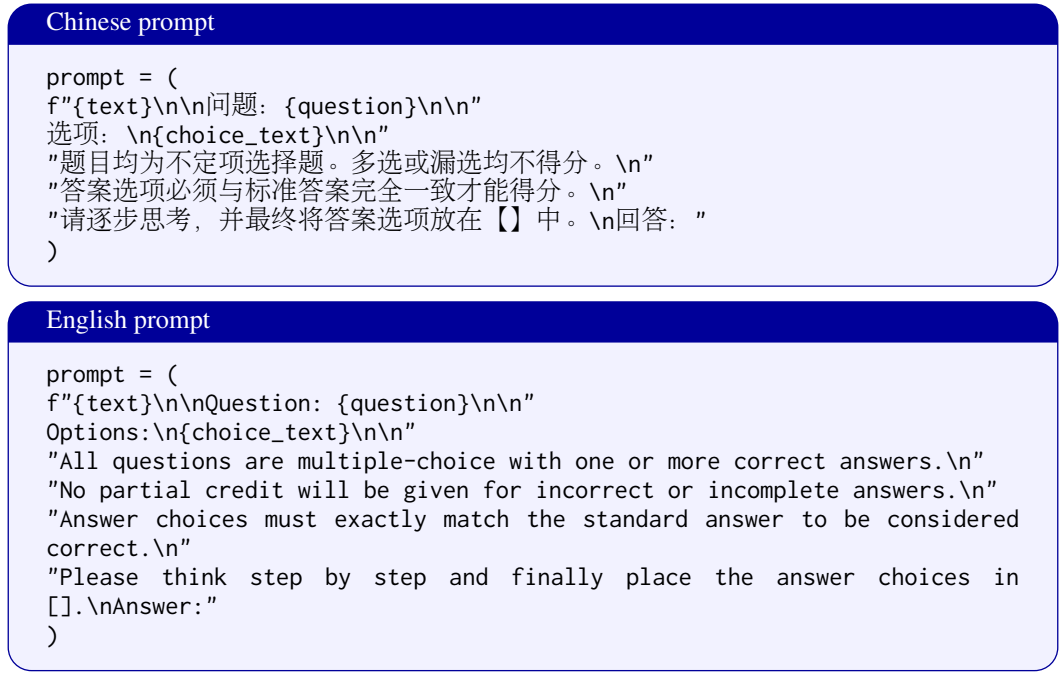

    \centering
    
\definecolor{lightgray}{gray}{0.9}

\begin{tcolorbox}[colframe=blue!60!black, colback=blue!5, arc=3mm, boxrule=0.5pt, title = Chinese prompt]
\texttt{prompt = (}\\
\texttt{f"\{text\}\textbackslash n\textbackslash n问题：\{question\}\textbackslash n\textbackslash n"}\\
\texttt{选项：\textbackslash n\{choice\_text\}\textbackslash n\textbackslash n"}\\
\texttt{"题目均为不定项选择题。多选或漏选均不得分。\textbackslash n"}\\
\texttt{"答案选项必须与标准答案完全一致才能得分。\textbackslash n"}\\
\texttt{"请逐步思考，并最终将答案选项放在【】中。\textbackslash n回答："}\\
\texttt{)}
\end{tcolorbox}

\begin{tcolorbox}[colframe=blue!60!black, colback=blue!5, arc=3mm, boxrule=0.5pt,  title = English prompt]
\texttt{prompt = (}\\
\texttt{f"\{text\}\textbackslash n\textbackslash nQuestion: \{question\}\textbackslash n\textbackslash n"}\\
\texttt{Options:\textbackslash n\{choice\_text\}\textbackslash n\textbackslash n"}\\
\texttt{"All questions are multiple-choice with one or more correct answers.\textbackslash n"}\\
\texttt{"No partial credit will be given for incorrect or incomplete answers.\textbackslash n"}\\
\texttt{"Answer choices must exactly match the standard answer to be considered correct.\textbackslash n"}\\
\texttt{"Please think step by step and finally place the answer choices in [].\textbackslash nAnswer:"}\\
\texttt{)}
\end{tcolorbox}
\caption{The English and Chinese prompts used in the evaluations.}
\label{fig:prompts}
\end{figure}

All models are tested using their official APIs. For models that allowed temperature adjustment, we set the temperature to 0.7. The English and Chinese prompts used in the evaluations are shown in Figure~\ref{fig:prompts}.

\section{Answer Extractor}
\label{app:answer_extractor}
In this test, we extract model-generated answers using a rule-based method. Since all questions are multiple-choice and we do not adopt the ICL~\citep{iclsurvey} approach. As a result, despite explicitly specifying the answer format in the prompt, some responses could not be extracted. However, the percentage of failed extractions is very low(which is shown in Table \ref{tab:extraction_failure_rate}) and do not impact overall performance. In the future, we may consider using a LLM as the answer extractor to improve accuracy, though this would come at a higher computational cost. We also observe that some models experienced unexplained interruptions when calling the API, which may be related to issues with the website links. Given sufficient resources, we plan to conduct further tests in the future to ensure the stability and reliability of the evaluation results.

\begin{table}[h!]
    \centering
    \caption{Unsuccessful extraction rates of different models}
    \begin{tabular}{l c}
        \toprule
Model Name & Unextracted Rate \\
\midrule
o1-mini & 0.0007 \\
deepseek-chat & 0.0387 \\
gpt-4o & 0.0341 \\
deepseek-r1-distill-qwen-32b & 0.0244 \\
claude-3-5-sonnet-20241022 & 0.0083 \\
o1-preview & 0.0156 \\
qwq-32B & 0.0463 \\
qwen-25-72B & 0.0074 \\
qwen-max & 0.0305 \\
glm-zero-preview & 0.0059 \\
o3-mini & 0.0126 \\
deepseek-reasoner & 0.0056 \\  
glm-4-plus & 0.0017 \\
\midrule
Overall Average & 0.0176 \\
\bottomrule
    \end{tabular}
    \label{tab:extraction_failure_rate}
\end{table}

\section{Human Evaluation}
\label{app:human_evaluation}

We recruited a total of 50 undergraduate students from Peking University to participate in the human evaluation, with 10 participants randomly assigned to each commonsense domain. Each domain consisted of 60 test questions. Except for the social relationship domain, which included 60 questions in Chinese, all other domains consisted of 30 questions in Chinese and 30 in English. To help participants stay focused during the evaluation, we divided the 60 questions into four online questionnaires, each containing 15 questions. Participants were required to complete each questionnaire within 2 hours. We released one questionnaire every two days, and the evaluation lasted 8 days. Participants are allowed to use pen and paper during the evaluation. The instruction presented to the participants was similar to the prompts given to LLMs, as presented in figure \ref{fig:prompts}. The entire human evaluation cost approximately \$10k.

After human evaluation, we calculated the accuracy of each participant in the 60 questions. For each domain, we computed the average accuracy and standard deviation of the 10 participants. Any score falling below the threshold of “average accuracy minus two standard deviation” was considered an outlier and excluded. In total, we collected 39 valid questionnaires. The number of valid results for each domain is reported in table \ref{tab:human_valid}.

\begin{table}[h] 
  \centering
  \caption{The number of valid human evaluation results per commonsense domain}
      \begin{tabular}{c c c c c | c}
        \toprule
        \textbf{Space} & \textbf{Time} & \textbf{Nature} & \textbf{Social} & \textbf{Mix} & \textbf{Total}\\
        \midrule
        7 & 8 & 9 & 8 & 7 & 39\\
        \bottomrule
      \end{tabular}
  \label{tab:human_valid}
\end{table}

\paragraph{Social Domain}
As shown in table \ref{tab:human_valid_social}, we found that human participants perform poorly on the questions of the hard level in social domain. The questions in the easy level and medium level cover general social relationships, where interpersonal connections are typically sparser and do not require relational calculations, in contrast to the Chinese family relationships in the hard-level questions. For example, a 'father's colleague' in a general social context lacks a single specific term, while a 'father's father' in a family context is simply 'grandfather'. This distinction makes family relationship questions inherently more difficult than those about general social relationships. Moreover, hard-level questions involve a larger number of entities, which further increases the difficulty for human participants to construct complex relationship networks (such as family networks).

\begin{table}[h] 
  \centering
  \caption{Human accuracy on 3 levels in social domain}
      \begin{tabular}{c c}
        \toprule
        Level & Human Test Accuracy \\
        \midrule
        Easy & $0.9857$ \\
        Medium & $0.9625$ \\
        Hard & $0.4458$ \\
        \bottomrule
      \end{tabular}
  \label{tab:human_valid_social}
\end{table}

\paragraph{Time Domain}
Due to the potential ambiguity of the temporal expression in language, it is difficult for human to understand the temporal relationship of the events. For instance, given the statements "There is a 1-day gap between Jack reads research papers and he cleans his dormitory room" and "Jack read research papers on Thursday", some may think that Jack cleans his dormitory room on Friday, while others may think Saturday. This ambiguity affects the human's understanding of the temporal relationship between events.

4 types of statements are questioned in time domain (see the "Imprecision" section in Appendix \ref{app:case_study}). The accuracy of human understanding on each type of statement is shown in Table \ref{tab:time_domain_human_analysis} below. We found that the Arith. questions are the most difficult for human to answer, since they are the most vague statements. Besides, Seq. \& Arith. questions are also ambiguious, leading to humans' misunderstanding.

\begin{table}[h!]
\centering
\caption{Human accuracy on 4 types of statements in time domain. Human has low accuracy on Seq. \& Arith. and Arith. statements, which are ambiguous.}
    \begin{tabular}{ccccc}
        \toprule
        \textbf{Language} & \textbf{Time.} & \textbf{Seq.} & \textbf{Arith.} & \textbf{Seq. \& Arith.} \\ \midrule
        cn & 1.00 & 0.85 & 0.49 & 0.71 \\
        en & 1.00 & 0.60 & 0.62 & 0.73 \\
        \bottomrule
    \end{tabular}
    
    \label{tab:time_domain_human_analysis}
\end{table}

\section{Detailed Performance on Different Domains}
\label{app:detail_performance}
\begin{figure*}
    \centering
    \includegraphics[width=1\linewidth]{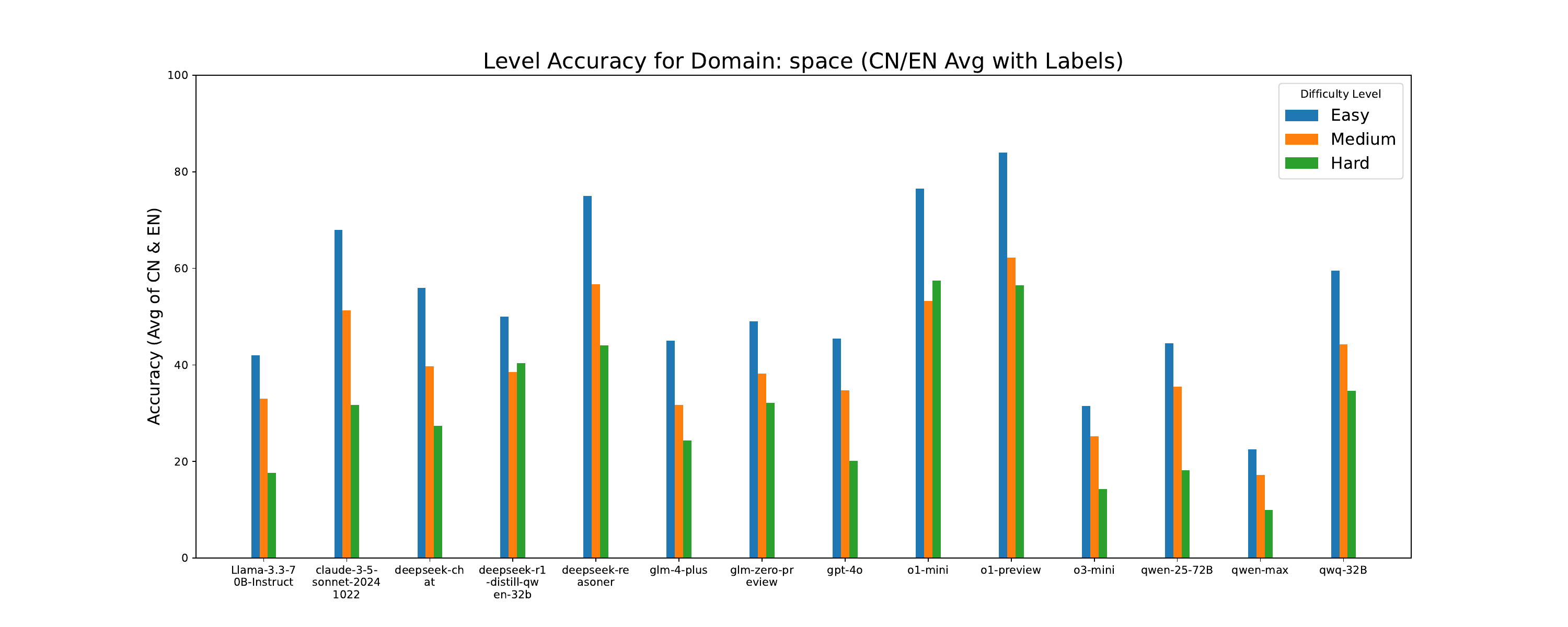}
    \caption{Level Accuracy on Space Domain}
    \label{fig:spacelevel}
\end{figure*}

\begin{figure*}
    \centering
    \includegraphics[width=1\linewidth]{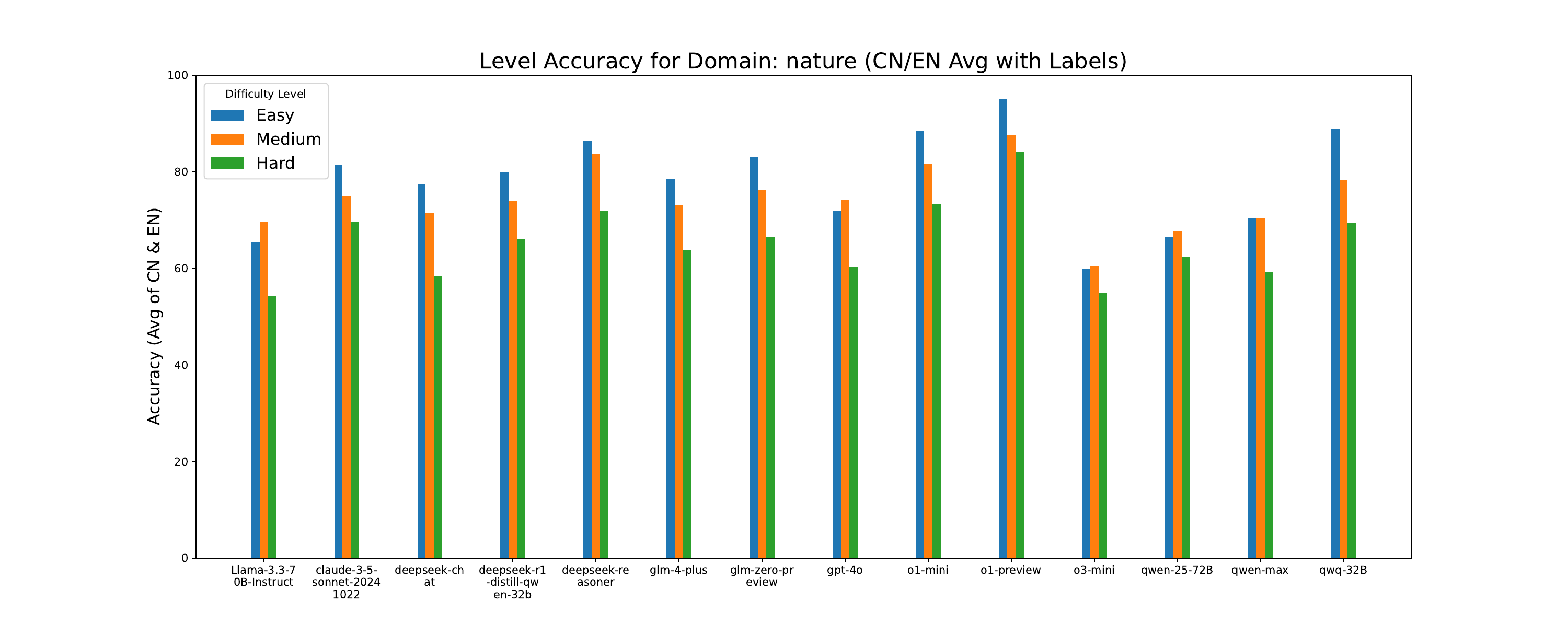}
    \caption{Level Accuracy on Nature Domain}
    \label{fig:naturelevel}
\end{figure*}

\begin{figure*}
    \centering
    \includegraphics[width=1\linewidth]{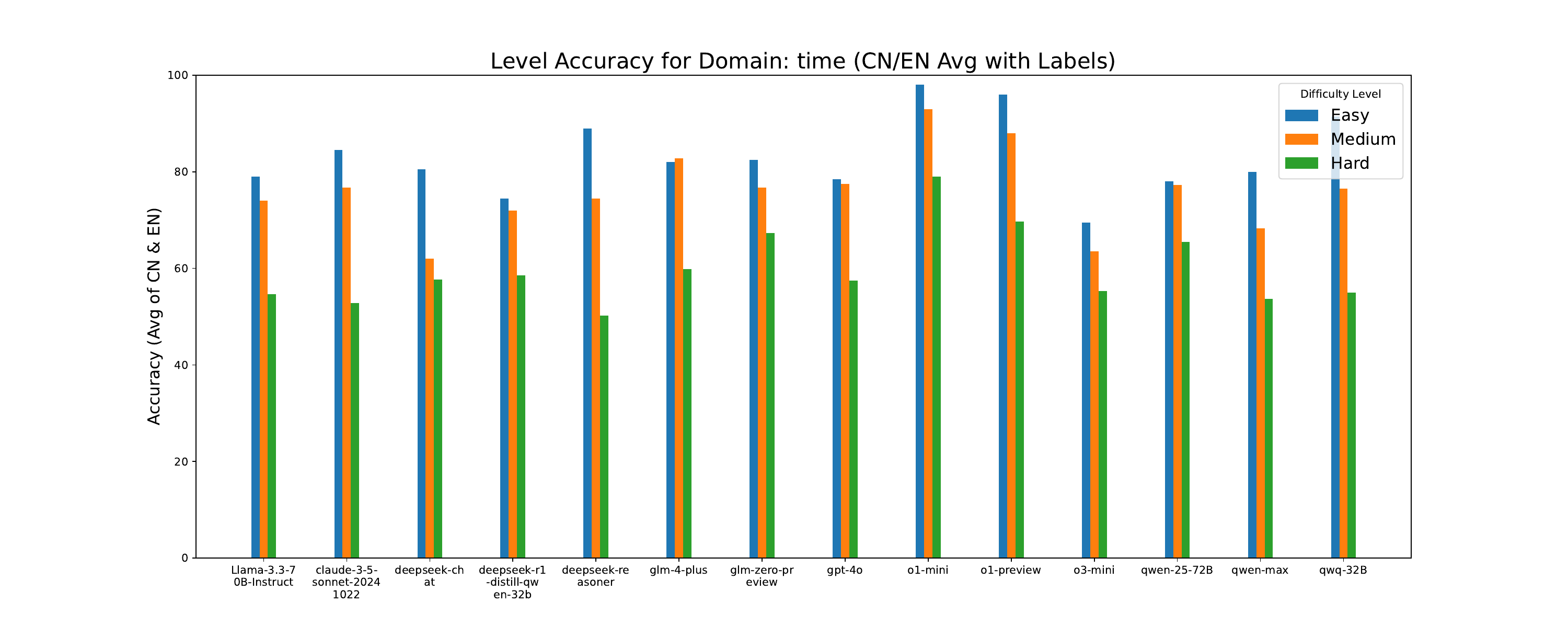}
    \caption{Level Accuracy on Time Domain}
    \label{fig:timelevel}
\end{figure*}

\begin{figure*}
    \centering
    \includegraphics[width=1\linewidth]{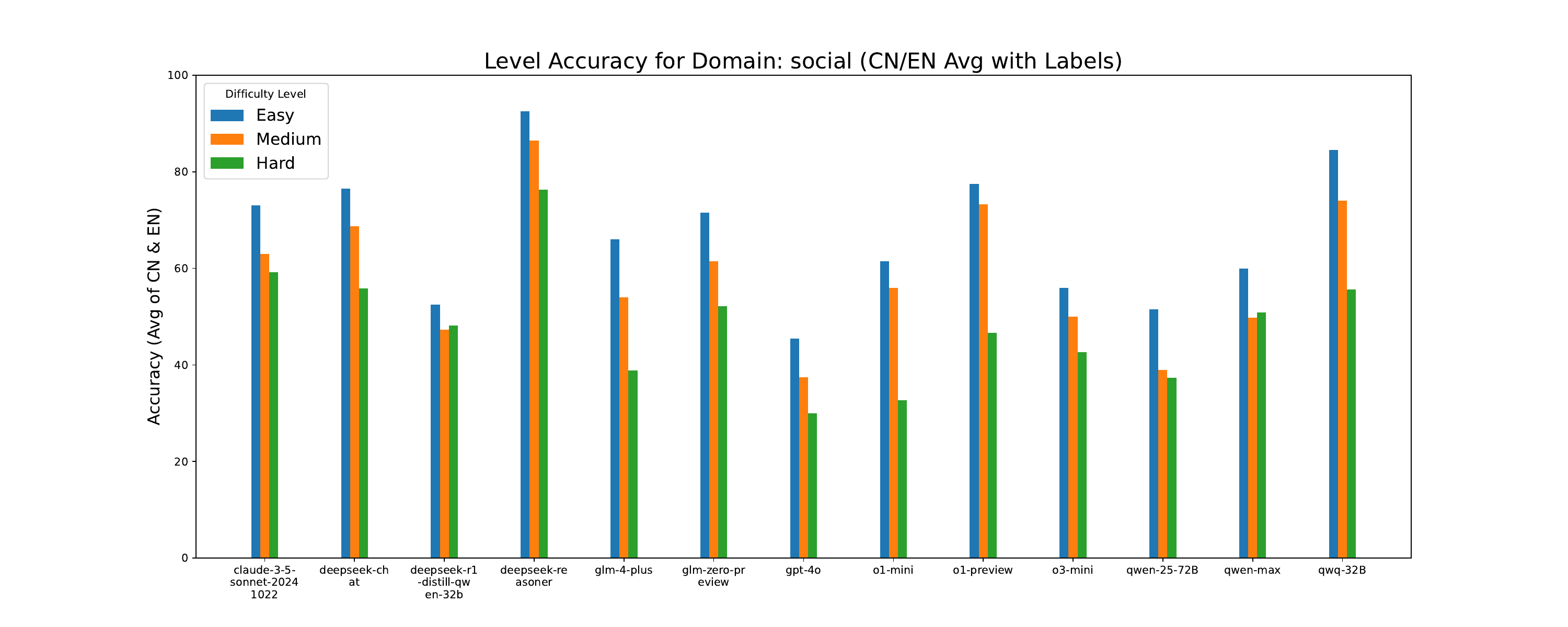}
    \caption{Level Accuracy on Social Domain}
    \label{fig:sociallevel}
\end{figure*}

\begin{figure*}
    \centering
    \includegraphics[width=1\linewidth]{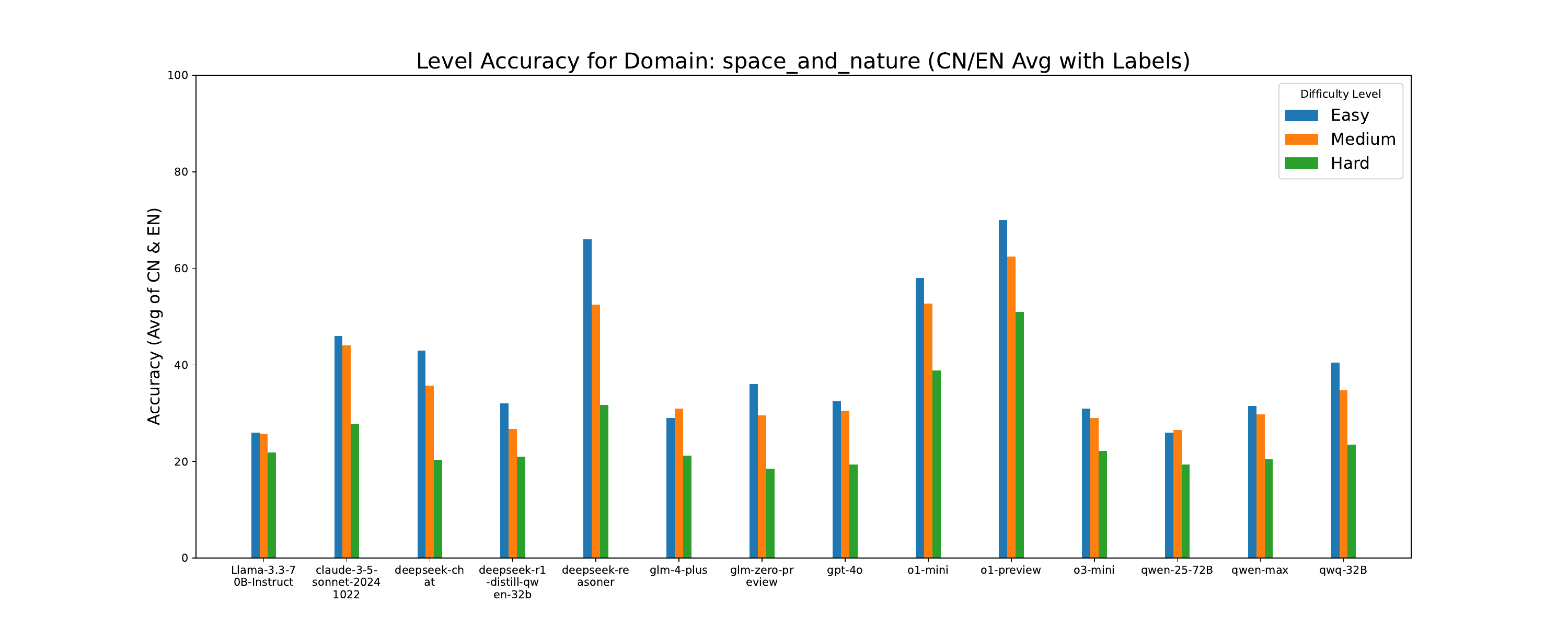}
    \caption{Level Accuracy on Mix Domain}
    \label{fig:mixlevel}
\end{figure*}

In this section, we provide a detailed analysis of model performance across different domains and difficulty levels. 
As illustrated in Figures~\ref{fig:spacelevel}, \ref{fig:naturelevel}, \ref{fig:timelevel}, \ref{fig:sociallevel}, and \ref{fig:mixlevel}, our difficulty classification system effectively differentiates question difficulty across most models, demonstrating its reliability in assessing model capabilities.

\begin{figure*}
  \centering 

  \includegraphics[width=0.36\linewidth]{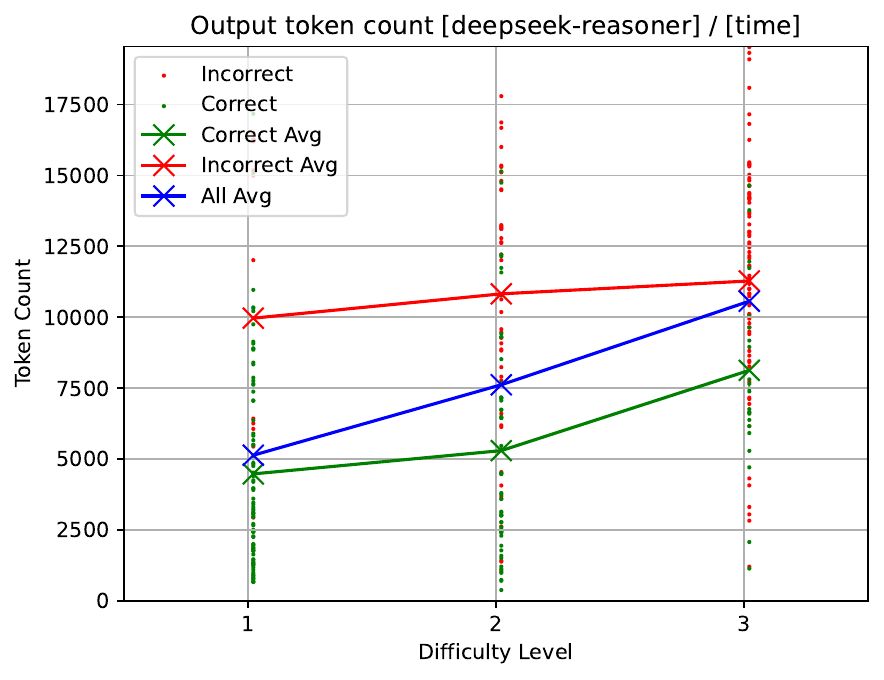}\hfill
  \includegraphics[width=0.36\linewidth]{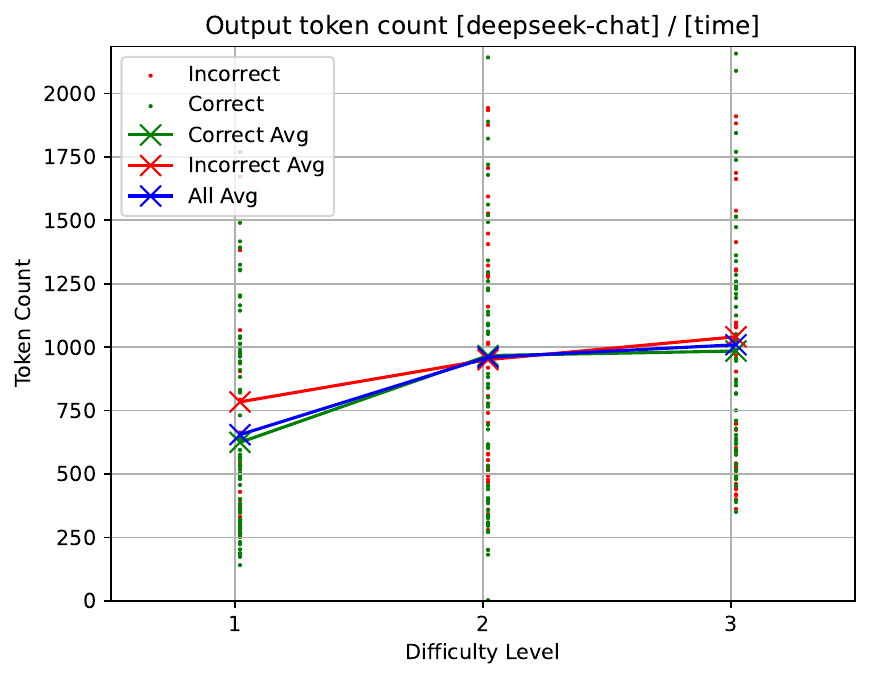}

  \includegraphics[width=0.36\linewidth]{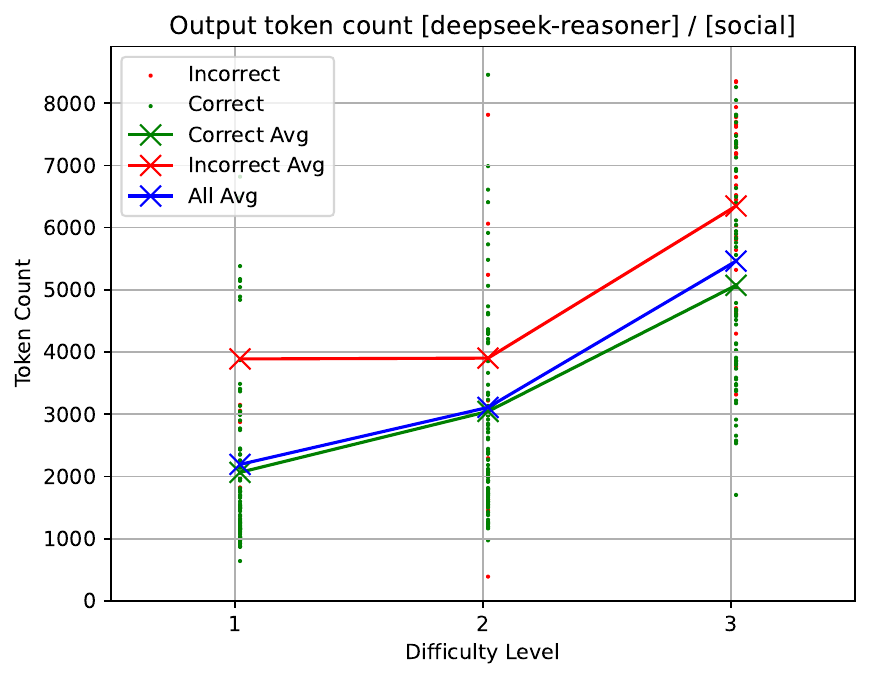}\hfill
  \includegraphics[width=0.36\linewidth]{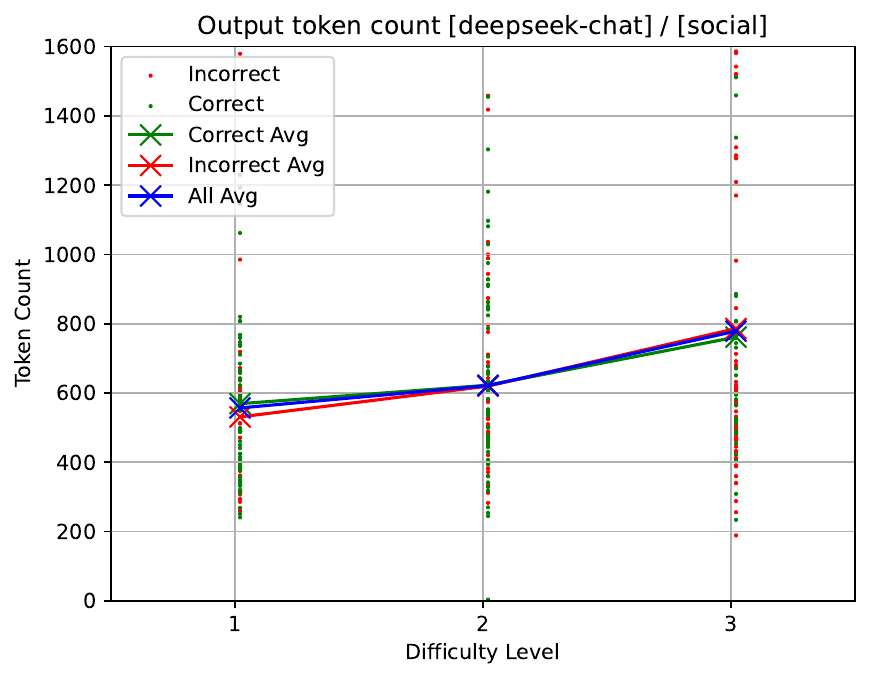}

  \includegraphics[width=0.36\linewidth]{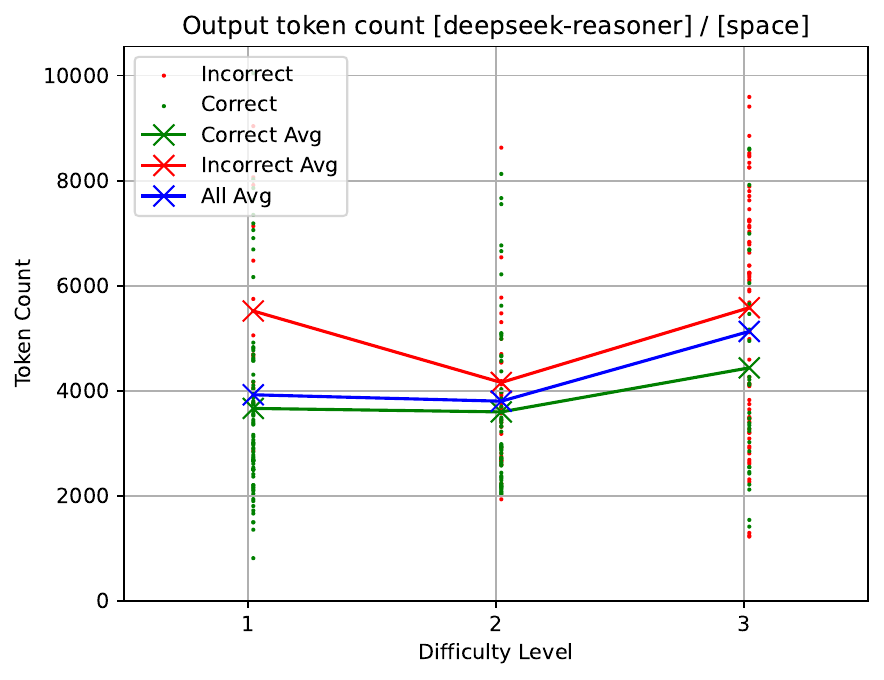}\hfill
  \includegraphics[width=0.36\linewidth]{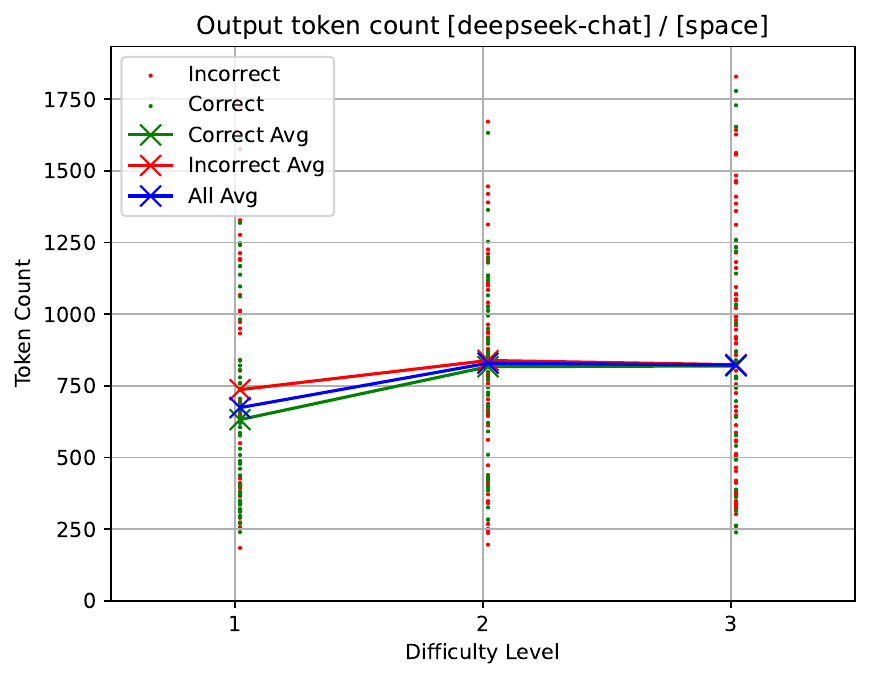}

  \includegraphics[width=0.36\linewidth]{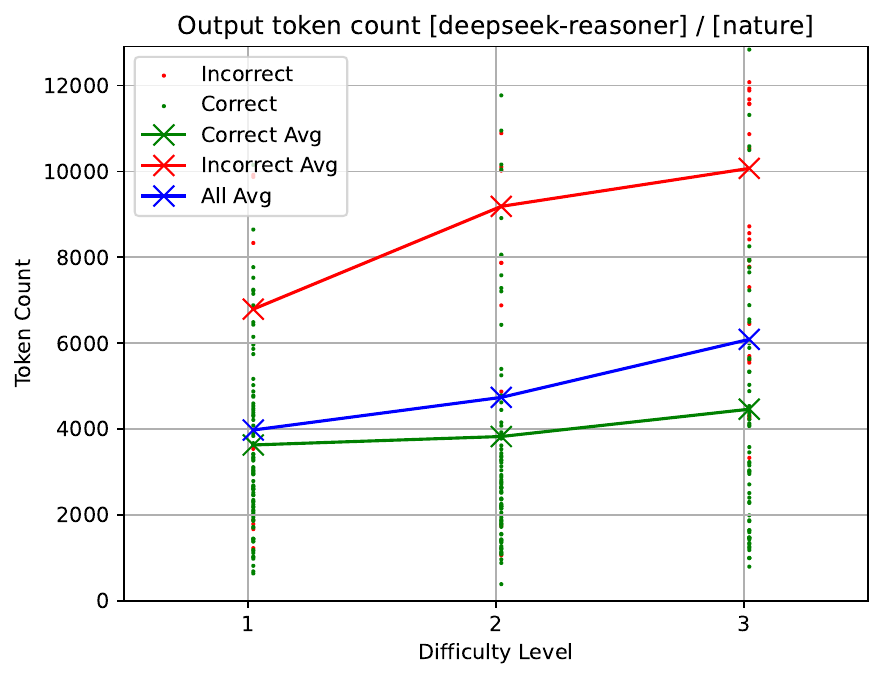}\hfill
  \includegraphics[width=0.36\linewidth]{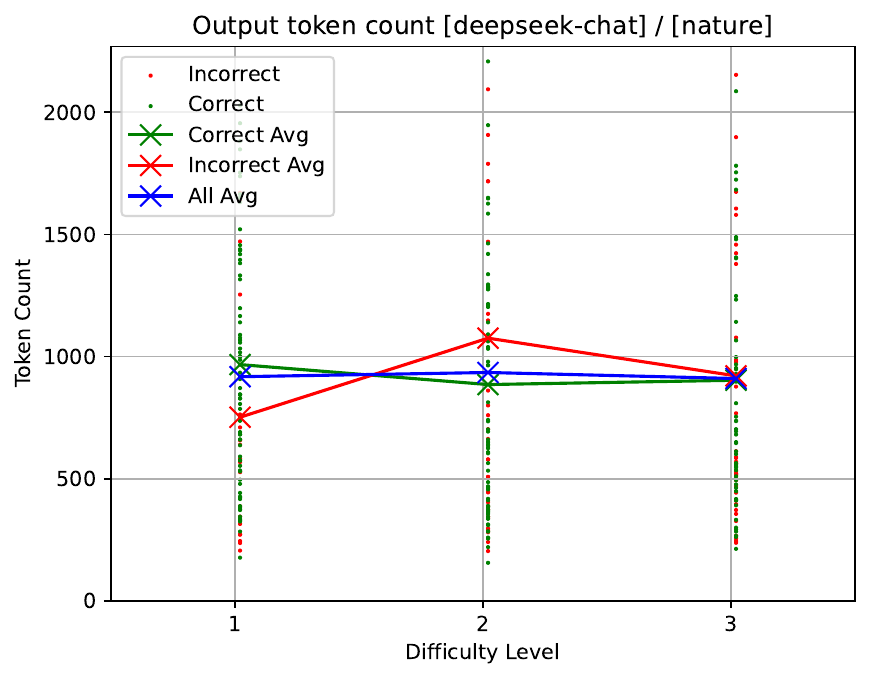}

  \includegraphics[width=0.36\linewidth]{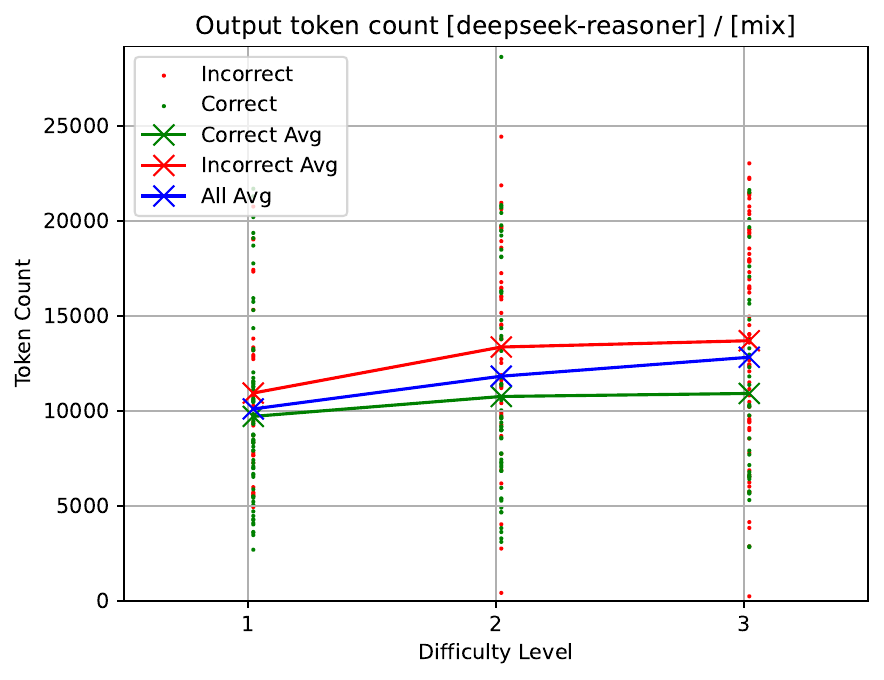}\hfill
  \includegraphics[width=0.36\linewidth]{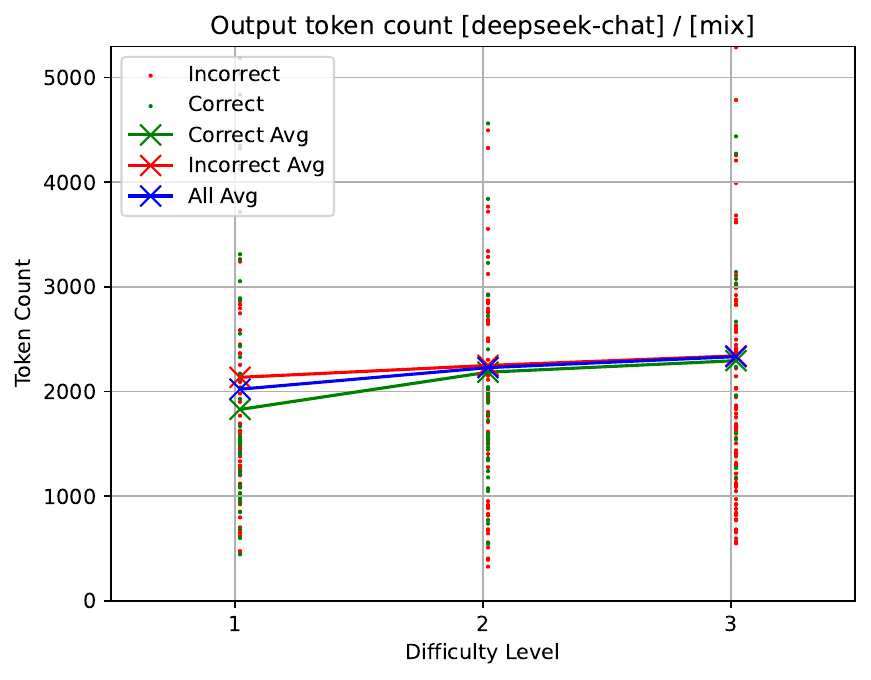}

  \caption{Relationship between response token count and question difficulty level. Left column: Results from deepseek-reasoner; Right column: Results from deepseek-chat. From top to bottom: Time, Social, Space, Nature, and Mix datasets (X-axis: Difficulty level, Y-axis: Token count).}
  \label{fig:token_num_statistics}
\end{figure*}

\section{Analysis of the Relationship Between Token Count and Question Difficulty}
\label{app:token_count}

Intuitively, models with reasoning capabilities typically require longer reasoning chains for more challenging questions, resulting in increased token counts in their outputs. Our experimental data corroborates this hypothesis and further reveals that reasoning-specialized models exhibit a stronger correlation between question difficulty and reasoning length compared to general models.

We conducted experiments using both deepseek-reasoner and deepseek-chat models across datasets of varying difficulty levels, collecting response metadata including token length and answer correctness. For each dataset and model combination, we generated scatter plots illustrating token length versus question difficulty level, supplemented with line plots demonstrating the relationships between difficulty levels and three key metrics: average token count, average token count for correct responses, and average token count for incorrect responses. The result is shown in Figure \ref{fig:token_num_statistics}.

The graphical analysis yields three principal findings:

(1) In deepseek-reasoner, within the same dataset, higher difficulty levels correspond to increased token counts in model outputs, indicating extended reasoning chains for more complex questions.

(2) The correlation between token count and difficulty level appears less pronounced in deepseek-chat, suggesting that reasoning-optimized models demonstrate superior capability in dynamically adjusting their cognitive processes according to the question complexity.

(3) Notably, in deepseek-reasoner, correct responses exhibit shorter average token lengths compared to incorrect ones within the same difficulty level. We hypothesize this phenomenon arises from two factors: (a) inherent variability in question difficulty within each difficulty level, and (b) an observable trend where more challenging questions (requiring longer reasoning chains) demonstrate higher error probabilities despite increased processing effort.

\section{Case Study}
\label{app:case_study}
\newtcolorbox[number within=section]{mybox}[2][]{colback=gray!5!white, colframe=black, coltitle=black, fonttitle=\bfseries, title=#2, #1, boxrule=0.5mm, left=2mm, right=2mm, top=2mm, bottom=2mm}

\subsection{Commonsense Error} 
\subsubsection{Entity Property Error} 


\begin{figure}[h!]
\centering
\begin{mybox}[title={}]
\tiny 
\fontsize{7}{5}\selectfont \textbf{\colorbox{gray!30}{Question-EN}} \\
The four enclosures in the zoo keep four different kinds of animals: cat, mandarin fish, tortoise, snuffer. Now we know that: \\The animal in enclosure No.2 and the animal in enclosure No.3 have 8 legs in total; \\The animal in enclosure No.3 has a shell; \\The animal in enclosure No.1 lives in sea water. \\
\_\_\_\_ is kept in enclosure No.4.\\
\colorbox{gray!30}{Question-CN} \\
动物园里的四个场馆分别养着猫、桂鱼、王八、海豚四种动物。已知：\\2号场馆中的动物与3号场馆中的动物加起来有8条腿；\\3号场馆中的动物有壳；\\1号场馆中的动物生活在海水中。\\
4号场馆中养的是\_\_\_\_。\\
\textbf{\colorbox{gray!30}{Options-EN}} \\
A. cat \\
B. mandarin fish \\
C. tortoise \\
D. snuffer \\
\textbf{\colorbox{gray!30}{Options-CN}} \\
A. 猫 \\
B. 桂鱼 \\
C. 王八 \\
D. 海豚 \\
\textbf{\colorbox{gray!30}{Correct Answer}} \\
B \\
\textbf{\colorbox{gray!30}{Model Name}} \\
o1-preview \\
\textbf{\colorbox{gray!30}{Model Answer-EN}} \\
 Let's analyze the given information step by step.\\
  Animals and Their Characteristics:\\
  ... [text omitted]...\\
  Mandarin Fish   - Legs: 0   - Shell: No   \textcolor{red}{- Habitat: Lives in seawater}\\
  ... [text omitted]...\\
  Answer: [D] \\
  \textbf{\colorbox{gray!30}{Model Answer-CN}} \\
 根据题目提供的条件，我们一步一步推理：\\
 (According to the information in the question, we analyze step by step) \\
  分析每个动物的特征：\\
  (Analyze the characteristics of animals)\\
  ... [text omitted]...\\
  桂鱼：没有腿，没有壳，\textcolor{darkgreen}{生活在淡水。}\\
  (Mandarin fish: no legs, no shell, \textcolor{darkgreen}{lives in freshwater.})\\
  ... [text omitted]...\\
  答案：【B】 \\
  (Answer:[B])\\
\end{mybox}
\caption{Example of entity property error in the model's response.}
\label{fig:attribute-error}
\end{figure}

We found that the model still lacks understanding of some low-frequency entity properties. As shown in Figure \ref{fig:attribute-error}, in the English question, the model mistakenly classified the mandarin fish (a freshwater fish) as a saltwater fish, leading to subsequent reasoning errors.

It is worth noting that such entity property errors are language-dependent. When the same question was asked in Chinese, the model correctly identified the mandarin fish as a freshwater fish and made the correct inference. We speculate that this is related to the frequency with which the entity appears in different language corpora. The mandarin fish is primarily found in East Asia and North Asia, where it is more frequently mentioned in Chinese corpora, while it is rarer in regions where English is the primary language. The lower occurrence frequency in English-language corpora likely leads to the model's misunderstanding of the mandarin fish's properties in an English context.

\subsubsection{Entity Relation Error} 
\begin{figure}
    \centering
    \includegraphics[width=\linewidth]{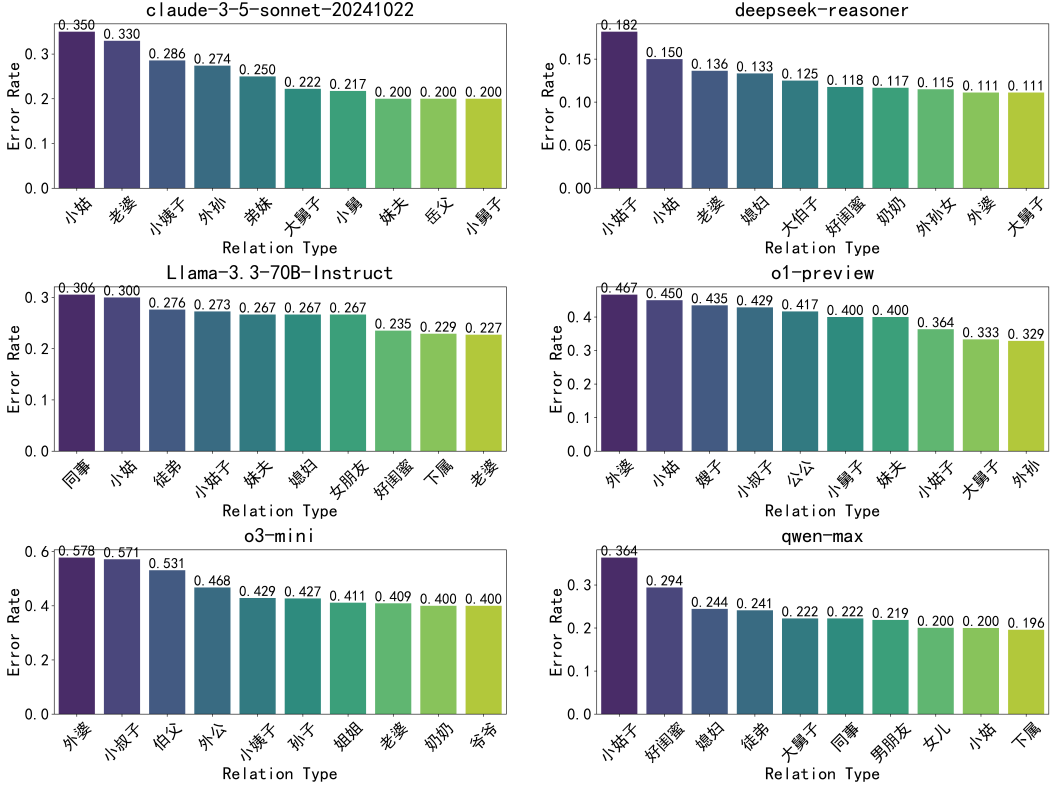}
    \caption{A Statistical Overview of Top 10 High-Frequency Error Terms in Large Language Model Performance on Social Relationship Comprehension}
    \label{fig:social-error-analysis}
\end{figure}
\begin{figure}[h!]
\centering
\begin{mybox}[title={}]
\tiny 
\fontsize{7}{5}\selectfont \textbf{\colorbox{gray!30}{Question}} \\
已知：李晓静是吴强的前妻，也是赵伟的前女友。孙大伟是吴强的挚友，也是钱静的老公。孙大伟是李晓静的领导。赵伟是钱静的同级生。以下选项正确的是\_\_\_ \\
(Given: Li Xiaojing is Wu Qiang's ex-wife and also Zhao Wei's ex-girlfriend. Sun Dawei is Wu Qiang's close friend and also Qian Jing's husband. Sun Dawei is Li Xiaojing's supervisor. Zhao Wei is Qian Jing's fellow student. The correct option among the following is \_\_\_.)\\
\textbf{\colorbox{gray!30}{Options}} \\
A. 李晓静的前男友是孙大伟的老婆的同事 \\
(A. Li Xiaojing's ex-boyfriend is the colleague of Sun Dawei's wife.)\\
B. 钱静的老公是赵伟的前女友的徒弟 \\
(B. Qian Jing's husband is the apprentice of Zhao Wei's ex-girlfriend.)\\
C. 赵伟的前女友是孙大伟的挚友的前妻 \\
(C. Zhao Wei's ex-girlfriend is the ex-wife of Sun Dawei's close friend.)\\
D. 赵伟的同级生是李晓静的领导的女朋友 \\
(D. Zhao Wei's fellow student is the girlfriend of Li Xiaojing's supervisor.)\\
\textbf{\colorbox{gray!30}{Correct Answer}} \\
C \\
\textbf{\colorbox{gray!30}{Model Name}} \\
GPT-4o \\
\textbf{\colorbox{gray!30}{Model Answer}} \\
我们来逐一分析每个选项：\\
(Let us analyze each option one by one:)\\
... [text omitted]...\\
A: 李晓静的前男友是孙大伟的老婆的同事。\\
(A. Li Xiaojing's ex-boyfriend is the colleague of Sun Dawei's wife.)\\
\textcolor{darkgreen}{- 李晓静的前男友是赵伟。\\
- 孙大伟的老婆是钱静。\\
- 钱静是赵伟的同级生，}\textcolor{red}{因此也是赵伟的同事。\\
- 所以，李晓静的前男友（赵伟）是孙大伟的老婆（钱静）的同事。\\
- 该选项正确。}\\
(\textcolor{darkgreen}{- Li Xiaojing's ex-boyfriend is Zhao Wei.\\
- Sun Dawei's wife is Qian Jing.\\
- Qian Jing is Zhao Wei's classmate,}\textcolor{red}{Therefore, he is also Zhao Wei's colleague.  \\
- Thus, Li Xiaojing's ex-boyfriend (Zhao Wei) is a colleague of Sun Dawei's wife (Qian Jing).  \\
- This option is correct.})\\
... [text omitted]...\\
所以，正确的答案是【\textcolor{darkgreen}{A}, \textcolor{red}{C}】。\\
(So, the correct answers are 【\textcolor{darkgreen}{A}, \textcolor{red}{C}】.)\\
\end{mybox}
\caption{Example of confusing different relationship in the model's response.}
\label{fig:social-error-analysis-1}
\end{figure}

\begin{figure}[h!]
\centering
\begin{mybox}[title={}]
\tiny 
\fontsize{7}{5}\selectfont \textbf{\colorbox{gray!30}{Question}} \\
已知：吴强是李晓静的同级生，也是赵伟的挚友。孙大伟是吴强的数位挚友中的一位，也是钱静的前男友。孙大伟是李晓静的师父。赵伟是钱静的领导。以下选项正确的是\_\_\_ \\
(Given: Wu Qiang is a fellow student of Li Xiaojing and also a close friend of Zhao Wei. Sun Dawei is one of Wu Qiang's several close friends and the ex-boyfriend of Qian Jing. Sun Dawei is Li Xiaojing's mentor. Zhao Wei is Qian Jing's supervisor. The correct option among the following is \_\_\_.)\\
\textbf{\colorbox{gray!30}{Options}} \\
A. 钱静的领导是孙大伟的挚友的哥哥 \\
(A. Qian Jing's supervisor is the elder brother of Sun Dawei's close friend.)\\
B. 赵伟的下属是李晓静的师父的前女友 \\
(B. Zhao Wei's subordinate is the ex-girlfriend of Li Xiaojing's mentor.)\\
C. 钱静的前男友是吴强的同级生的徒弟 \\
(C. Qian Jing's ex-boyfriend is the apprentice of Wu Qiang's fellow student.)\\
D. 吴强的某位名字中带有伟的挚友是孙大伟的前女友的下属 \\
(D. One of Wu Qiang's close friends whose name contains the character "Wei" is the subordinate of Sun Dawei's ex-girlfriend.)\\
\textbf{\colorbox{gray!30}{Correct Answer}} \\
B \\
\textbf{\colorbox{gray!30}{Model Name}} \\
GPT-4o \\
\textbf{\colorbox{gray!30}{Model Answer}} \\
为了确定正确的答案，我们需要一步一步地进行逻辑推理，理清各个角色及其关系。\\
(To determine the correct answer, we need to proceed with logical reasoning step by step, clarifying the roles and relationships of each individual.)\\
... [text omitted]...\\
C: 钱静的前男友是吴强的同级生的徒弟\\
(C. Qian Jing's ex-boyfriend is the apprentice of Wu Qiang's fellow student.)\\
\textcolor{darkgreen}{根据信息，钱静的前男友是孙大伟，而孙大伟同时也是李晓静的师父，因此孙大伟是吴强的同级生（李晓静）的师父}\textcolor{red}{（徒弟），此项正确。}\\
(\textcolor{darkgreen}{According to the information, Qian Jing's ex-boyfriend is Sun Dawei, who is also Li Xiaojing's mentor. Therefore, Sun Dawei is the mentor of Wu Qiang's fellow student (Li Xiaojing)}\textcolor{red}{(apprentice), this item is correct.})\\
... [text omitted]...\\
因此，正确的选项是【\textcolor{darkgreen}{B}, \textcolor{red}{C}】。\\
(Therefore, the correct options are 【\textcolor{darkgreen}{B}, \textcolor{red}{C}】.)\\
\end{mybox}
\caption{Example of misinterpreting asymmetric relationships in the model's response.}
\label{fig:social-error-analysis-2}
\end{figure}


We find that the model often confuses closely related social relationships. As shown in Figure \ref{fig:social-error-analysis-1}, the model correctly handles most reasoning steps, but it treats "classmate" and "colleague" as the same, confusing these relationships. It also confuses "mentor" and "leader," as well as "ex-husband" and "ex-boyfriend."

Additionally, the model misinterprets asymmetric relationships. As shown in Figure \ref{fig:social-error-analysis-2}, it treats "A is B's mentor" and "A is B's apprentice" as identical, failing to understand the mentor-apprentice relationship. Similar errors occur with "A is B's subordinate" and "B is A's subordinate."

To further investigate the models' comprehension of social relationships, we compiled statistics on the frequency of errors made by the models on test questions pertaining to different social relations. As shown in Figure \ref{fig:social-error-analysis}, Current Large Language Models exhibit a general proclivity for error when processing social relationship terminologies, particularly with affinal (in-law) relationships, generational terms (especially those requiring distinctions between maternal/paternal lines, relative age, and gender), and certain everyday yet ambiguously-bounded non-consanguineous relationships (e.g., friends, colleagues). This predisposition towards error likely stems from inadequacies in the models' capacity to learn and infer the conventional naming systems, cultural contexts, and complex network structures inherent in human social organization. While error hotspots vary slightly among different models, possibly reflecting variations in their training datasets and algorithmic emphases, the overarching challenge remains.

\subsection{Reasoning Error} 
\subsubsection{Logical Contradiction} 
\begin{figure}[h!]
\centering
\begin{mybox}[title={}]
\tiny 
\fontsize{7}{5}\selectfont \textbf{\colorbox{gray!30}{Question}} \\
Flute, conch, fresh kidney bean, mess tin, duckweed, envelope, six items are placed on a three-tier shelf, which is positioned against the south wall of the store. Each tier has two sections, with one type of item placed in the east section and one in the west section. A customer is standing in front of the shelf. When describing the positional relationships of the items, it is agreed that the customer's own left and right will be used as a reference, with the east section being on the left and the west section being on the right. It is known that:\\
The envelope is located in the upper left corner of the flower or grass and the two are not separated by a layer,\\
The flower or grass and the animal with no legs are vertically adjacent to each other,\\
The musical instrument is on the same level as the vegetable,\\
The item with shell is on the left side of the bottom tier,\\
The right side neighbor of the musical instrument is above the upper right side of the envelope.\\
The flute is to the left of \_\_\_ \\
\textbf{\colorbox{gray!30}{Options}} \\
A. mess tin \\
B. conch \\
C. evnelope \\
D. None of the above \\
\textbf{\colorbox{gray!30}{Correct Answer}} \\
D \\
\textbf{\colorbox{gray!30}{Model Name}} \\
GPT-4o \\
\textbf{\colorbox{gray!30}{Model Answer}} \\
... [text omitted]... \\
5. The right side neighbor of the musical instrument is above the upper right side of the envelope.\\
 ... [text omitted]... \\
 \textcolor{darkgreen}{Since clue 5 states that the right neighbor of the flute (musical instrument) is above the envelope's upper right, this implies that they must be on different tiers, and the flute is above the envelope.}\\
  ... [text omitted]... \\
\textcolor{red}{Given clue 5, if envelope is on the top left, the musical instrument (flute) must be in the middle tier.}\\
  ... [text omitted]... \\
  Thus answering the question, "Flute is to the left of \_\_\_", 、\textcolor{darkgreen}{flute is to the left of the fresh kidney bean.} \\
  \textcolor{red}{Answer: [A: mess tin]}\\
\end{mybox}
\caption{Example of logical contradiction in the model's response.}
\label{fig:self-contrast}
\end{figure}
We notice that the model's response may sometimes be self-contrast. As shown in Figure~\ref{fig:self-contrast}, the model infers rightly that the flute is above the envelope based on clue 5. However, in the following analysis, the model assumes that the envelope is on the top tier and the flute is on the middle tier based on the same clue, which contrasts the previous analysis.
Additionally, the model also exhibit contradiction when summarizing the answer.
According to the model's analysis, the flute is to the left of the fresh kidney bean, but in the end, it still outputs the answer as "the flute is to the left of the mess tin".
This implies that LLMs have not yet possess logical reasoning abilities aligned with humans.

\subsubsection{Imprecision}
\begin{figure}[h!]
\centering
\begin{mybox}[title={}]
\tiny 
\fontsize{7}{5}\selectfont \textbf{\colorbox{gray!30}{Question}} \\
Jack is a college student, and here are his weekly plans: \\
(1)On Monday, Jack learns Japanese; \\
(2)On Wednesday, Jack plays badminton; \\
(3)Jack goes jogging 2 days after he learns Japanese; \\
(4)Jack cleans his dormitory room 1 day after he goes jogging; \\
(5)Jack practices the guitar 1 day after he cleans his dormitory room; \\
(6)1 day after Jack practices the guitar, he reads research papers.\\
5 days before \_\_\_\_, Jack plays badminton.\\
\textbf{\colorbox{gray!30}{Options}} \\
A.Jack learns Japanese\\
B.Jack practices the guitar\\
C.Jack reads research papers\\
D.Jack goes jogging\\
\textbf{\colorbox{gray!30}{Correct Answer}} \\
A \\
\textbf{\colorbox{gray!30}{Model Name}} \\
DeepSeek-Chat \\
\textbf{\colorbox{gray!30}{Model Answer}} \\
Let's break this down step by step:\\
1. \textbf{Monday}: Jack learns Japanese. \\ 
2. \textbf{Wednesday}: Jack plays badminton. \\
3. \textcolor{red}{\textbf{2 days after learning Japanese (Wednesday + 2 days = Friday)}: Jack goes jogging.}\textcolor{darkgreen}{(In fact, Jack learns Japanese on Monday, not Wednesday.)}\\
...[text omitted]...\\
Now, the question asks: "5 days before \_\_\_\_, Jack plays badminton."\\
Jack plays badminton on \textbf{Wednesday}.  \\
5 days before Wednesday is \textbf{Friday}.  \\
\textcolor{red}{From the schedule, we know that on Friday, Jack \textbf{goes jogging}.}  \\
\textcolor{red}{Thus, the correct option is: \textbf{D: Jack goes jogging}.}  \\
The answer is [\textcolor{red}{D}]. \\
\end{mybox}
\caption{Example of imprecision in the model's response.}
\label{fig:imprecision}
\end{figure}
 In the reasoning process, models may be affected by irrelevant information contained in previous outputs, leading to imprecision in final results.
    As shown in Figure \ref{fig:imprecision}, when calculating the time that "Jack goes jogging", the model uses the existing condition "2 days after learning Japanese". 
    However, it is misled by closer statements that are unrelated, since "Wednesday" in context is misused in the process. This imprecision results in an error in reasoning.\par
    In time domain, 4 types of statements are questioned:
    \begin{enumerate}
        \item Time of Events (Time.), such as "Jack met his future wife in 1920".
        \item Sequence of Events (Seq.), such as "Jack met his future wife after he graduated from high school".
        \item Arithmetic of Time (Arith.), such as "There is a 2-year gap between Jack graduated from high school and he met his future wife".
        \item Sequence of Events and Arithmetic of Time (Seq. \& Arith.), such as "Jack met his future wife 2 years after he graduated from high school".
    \end{enumerate}
    Since arithmetic is more difficult than sequence comparison in reasoning, it is easier to be misled by irrelevant information.
    Thus, accuracy of the model in Arith. questions is lower than that in Seq. questions, and models has the lowest accuracy in Seq. \& Arith. questions (shown in the table below).\par
    \begin{table*}[h!]
        \centering
        \caption{Accuracy of models on different time reasoning types. Since Arith. is harder than Seq., the accuracy of Arith. is lower than that of Seq.}
        \begin{tabular}{ccccccccc}
            \toprule
            \multirow{2}{*}{\textbf{Models}} & 
            \multicolumn{2}{c}{\textbf{Time.}} & 
            \multicolumn{2}{c}{\textbf{Seq.}} & \multicolumn{2}{c}{\textbf{Arith.}} & \multicolumn{2}{c}{\textbf{Seq. \& Arith}} \\ \cmidrule{2-9}
             & cn & en & cn & en & cn & en & cn & en \\ \midrule
            claude-3-5-sonnet & 0.67 & 0.74 & 0.65 & 0.82 & 0.58 & 0.70 & 0.46 & 0.58 \\
            deepseek-reasoner & 0.45 & 0.57 & 0.46 & 0.76 & 0.35 & 0.60 & 0.31 & 0.54 \\
            gpt-4o & 0.68 & 0.62 & 0.73 & 0.85 & 0.62 & 0.69 & 0.57 & 0.59 \\
            o1-mini & 0.88 & 0.75 & 0.96 & 0.95 & 0.84 & 0.79 & 0.81 & 0.79 \\
            o1-preview & 0.62 & 0.74 & 0.93 & 0.91 & 0.71 & 0.70 & 0.70 & 0.75 \\
            qwen-2.5-72B & 0.65 & 0.84 & 0.60 & 0.83 & 0.54 & 0.74 & 0.43 & 0.65 \\
            qwen-max & 0.51 & 0.70 & 0.60 & 0.83 & 0.54 & 0.74 & 0.43 & 0.65 \\
            \bottomrule
        \end{tabular}
        
    \end{table*}
    According to the text of questions, this imprecision could be cause by (1) a wrong understanding of the entity or (2) a wrong understanding of the property.
    Since the occurrence of this error is related to next token prediction, which is regarded as the cornerstone of LLMs, similiar examples can also be found in other fields.

\subsubsection{Inference Error} 

\begin{figure}[h!]
\centering
\begin{mybox}[title={}]
\tiny 
\fontsize{7}{5}\selectfont \textbf{\colorbox{gray!30}{Question}} \\
张天师(David)、曹国舅(Elizabeth)、张果老(Mary)、何仙姑(Jennifer)、汉钟离(William)、吕洞宾(Robert)六座神像在神坛中围成一个圆圈，每座神像都背对神坛中心。六个神像的位置恰好落在正六边形的六个顶点上。任意相邻两个神像之间的距离相等，大约为一米。已知：\\
张天师的左边起第五个是吕洞宾;\\
曹国舅在何仙姑左侧紧邻位置;\\
从吕洞宾的左边数起第三个位置是张果老;\\
吕洞宾在汉钟离右边数起第二个位置;\\
张果老在何仙姑右侧紧邻位置。\\
张天师紧邻\_\_\_。\\
(David, Elizabeth, Mary, Jennifer, William, Robert,  — these six statues are arranged in a circle on the altar, each facing away from the center of the altar. The positions of the six statues correspond precisely to the six vertices of a regular hexagon. The distance between any two adjacent statues is equal, approximately one meter. It is known that: \\
Robert is the fifth in sequence starting to the left of David;\\
Elizabeth is directly to the left of Jennifer;\\
Mary occupies the third position to the left of Robert;\\
Robert occupies the second position to the right of William;\\
Mary is directly to the right of Jennifer.\\
David is next to \_\_\_.)\\

\textbf{\colorbox{gray!30}{Options}} \\
A. 张果老(Mary) \\
B. 何仙姑(Jennifer) \\
C. 吕洞宾(Robert) \\
D. 曹国舅(Elizabeth) \\

\textbf{\colorbox{gray!30}{Correct Answer}} \\
B,C \\
\textbf{\colorbox{gray!30}{Model Name}} \\
GPT-4o \\
\textbf{\colorbox{gray!30}{Model Answer}} \\
  根据题目提供的信息，我们可以推理出神像的排列顺序：\\
  (Based on the information provided in the question, we can deduce the arrangement order of the statues:)\\
  \textcolor{red}{张天师的左边起第五个是吕洞宾：这意味着吕洞宾在张天师的对面。}\\
  \textcolor{red}{(Robert is the fifth in sequence starting to the left of David, This means that Robert is opposite David. )}\\
  \textcolor{darkgreen}{(In fact, in this scenario, the fifth position on the left is equivalent to the first position on the right, not directly opposite.)}\\
  ... [text omitted]...\\
  \textcolor{red}{所以，张天师紧邻张果老。答案是：A. 张果老}\\
  \textcolor{red}{(so, David is next to Mary. The answer is A. Mary)}
\end{mybox}
\caption{Example of inference error in the model's response.}
\label{fig:reasoning-error}
\end{figure}
As shown in figure \ref{fig:reasoning-error}, the question states that six entities are arranged in a hexagonal layout, creating a ring where the $n$-th position on the left is equivalent to the $(6-n)$-th position on the right. Considering the angular relationships of the hexagon's vertices, it is the third position on the left or the third on the right that is directly opposite. Such an error indicates that the model did not correctly understand the configuration of entities based on the scenario described in the question, leading to incorrect reasoning about their relative positions. This type of error are related to the model's ability to construct and understand scenario.

\subsection{Overthinking} 
\begin{figure}[h!]
\centering
\begin{mybox}[title={}]
\tiny 
\fontsize{7}{5}\selectfont \textbf{\colorbox{gray!30}{Question}} \\
On a wall pasted photos of four different items: leather jacket, carnation, birdcage, coriander. Now we know that: \\
The item on photo No.3 is not a clothing; \\
The item on photo No.2 is not a clothing; \\
The flower of the item on photo No.3 is white; \\
The item on photo No.4 is not a tool; \\
The item on photo No.4 is not a clothing; \\
The item on photo No.3 is not a flower or grass.\\
Carnation is on photo No.\_\_\_\_.\\
\textbf{\colorbox{gray!30}{Options}} \\
A. 1 \\
B. 2 \\
C. 3 \\
D. 4 \\
\textbf{\colorbox{gray!30}{Correct Answer}} \\
C \\
\textbf{\colorbox{gray!30}{Model-1 Name}} \\
o1-preview \\
\textbf{\colorbox{gray!30}{Model-1 Answer}} \\
Given Constraints:\\
... [text omitted]...\\
3. The flower of the item on photo No.3 is white.\\
... [text omitted]...\\
First, let's list out the possible categories for each item:\\
Birdcage: Tool\\
Coriander: Grass/Herb (has small white flowers)\\
 ... [text omitted]... \\
 \textcolor{red}{Perhaps the birdcage is decorated with white flowers or has a design featuring white flowers.}\\
 ... [text omitted]...\\
 \textcolor{red}{Therefore, the item on photo No.3 must be the birdcage}.\\
  ... [text omitted]...\\
  Answer: [B] \\
  \textbf{\colorbox{gray!30}{Model-2 Name}} \\
Llama3.3-70B-Instruct \\
\textbf{\colorbox{gray!30}{Model-2 Answer}} \\
given the items listed (leather jacket, carnation, birdcage, coriander), coriander has small white flowers, and it's not a clothing, tool, or exclusively a flower in the context of these clues\\
 ... [text omitted]...\\
The birdcage, not being clothing, could fit on No.2, given that No.3 has a specific description that doesn't fit the birdcage directly\\
 ... [text omitted]...\\
 The answer is [D]\\
\end{mybox}
\caption{Example of overthinking in the model's response.}
\label{fig:overthinking}
\end{figure}

We find that some reasoning-specialized LLMs, such as the o1 and r1 series, sometimes overthink and introduce special cases to fit the given conditions. As shown in Figure \ref{fig:overthinking}, the condition “The flower of the item on photo No.3 is white” clearly points to coriander in photo No.3. However, o1 suggests that a birdcage can be decorated with white flowers, making it a possible answer. In contrast, general-purpose models like qwen-max do not overthink and correctly answer the question. This implies that the training methods for reasoning-specialized models emphasizing logical consistency and depth of reasoning may lead them to miss more staightforword solutions.

\subsection{Answer Aggregation Error}
\begin{figure}[h]
\centering
\begin{mybox}[title={}]
\tiny 
\fontsize{7}{5}\selectfont \textbf{\colorbox{gray!30}{Question}} \\
Rolling luggage, eyebrow pencil, plaice, Chinese herbaceous peony, turkey, port wine, six items are placed on a three-tier shelf, which is positioned against the south wall of the store. Each tier has two sections, with one type of item placed in the east section and one in the west section. A customer is standing in front of the shelf. When describing the positional relationships of the items, it is agreed that the customer's own left and right will be used as a reference, with the east section being on the left and the west section being on the right. It is known that:\\The animal with no legs is directly to the eyebrow pencil's left,\\The rolling luggage and the eyebrow pencil are separated by one tier,\\The bird is on the first floor,\\The flower or grass is above the animal with 2 legs in a straight line,\\The right side neighbor of the animal with no legs is above the upper right side of the animal with 2 legs,\\The drink is somewhere to the right of the flower or grass,\\The bird is located at the bottom left of the eyebrow pencil and there is one tier separating the two,\\The rolling luggage and the drink are vertically adjacent to each other.\\
port wine and \_\_\_ are vertically adjacent to each other \\
\textbf{\colorbox{gray!30}{Options}} \\
A. eyebrow pencil \\
B. turkey \\
C. rolling luggage \\
D. None of the above \\
\textbf{\colorbox{gray!30}{Correct Answer}} \\
AC \\
\textbf{\colorbox{gray!30}{Model Name}} \\
o1-preview \\
\textbf{\colorbox{gray!30}{Model Answer}} \\
... [text omitted]... \\
\textcolor{darkgreen}{- Top Floor (3rd Tier):\\  - Left (East): Plaice (animal with no legs)  - Right (West): Eyebrow pencil\\- Middle Floor (2nd Tier):  - Left (East): Chinese herbaceous peony (flower)  - Right (West): Port wine (drink)\\- Bottom Floor (1st Tier):  - Left (East): Turkey (bird, animal with 2 legs)  - Right (West): Rolling luggage}\\
Answering the question:
\textcolor{darkgreen}{Port wine and rolling luggage are vertically adjacent to each other.} \textcolor{red}{So, the correct choice is C.\\**Answer:** [C]}\\
\end{mybox}
\caption{Example of answer aggregation error in the model's response.}
\label{fig:Answer-Aggregation-Error}
\end{figure}
Despite mentioning that our questions are multiple-choice in the prompt, the models still tend to treat them as single-choice questions. As shown in Figure~\ref{fig:Answer-Aggregation-Error}, the model outlines the right arrangement but stops checking once it finds one correct answer.
This may be due to the high frequency of single-choice questions in inference data, which leads the models to learn a shortcut outputting the answer once they find one answer. 

Additionally, logical contradiction may also contribute to answer aggregation errors. The model may analyse A is right but answer B, as shown in Figure~\ref{fig:self-contrast}.

\subsection{Analysis on different behaviour in Chinese and English}
\label{app:language}
It is generally believed that large language models behave better in English than in non-English languages due to the quality and quantity of training data~\cite{zhang-etal-2023-dont}. However, on the contrary, our results show that most models perform better on Chinese questions than on English questions in space, nature and mixed domains. We filtered and analyzed 100 questions with the greatest performance differences between Chinese and English, and found that these differences are primarily due to the following reasons:

\paragraph{Difference in the frequency of entities}
\begin{figure}[h!]
\centering
\begin{mybox}[title={}]
\tiny 
\fontsize{7}{5}\selectfont \textbf{\colorbox{gray!30}{Question-CN}}\\
动物园里的四个场馆分别养着猫、桂鱼、王八、海豚四种动物。已知：\\2号场馆中的动物与3号场馆中的动物加起来有8条腿；\\3号场馆中的动物有壳；\\1号场馆中的动物生活在海水中。\\
4号场馆中养的是\_\_\_\_。\\
\textbf{\colorbox{gray!30}{Options-CN}} \\
A. 猫 \\
B. 桂鱼 \\
C. 王八 \\
D. 海豚 \\
{Question-EN} \\
The four enclosures in the zoo keep four different kinds of animals: cat, mandarin fish, tortoise, snuffer. Now we know that: \\The animal in enclosure No.2 and the animal in enclosure No.3 have 8 legs in total; \\The animal in enclosure No.3 has a shell; \\The animal in enclosure No.1 lives in sea water. \\
\_\_\_\_ is kept in enclosure No.4.\\
\textbf{\colorbox{gray!30}{Options-EN}}\\
A. cat \\
B. mandarin fish \\
C. tortoise \\
D. snuffer \\
\textbf{\colorbox{gray!30}{Correct Answer}} \\
B \\
\textbf{\colorbox{gray!30}{Model Name}} \\
o1-preview \\
\textbf{\colorbox{gray!30}{Model Answer-CN}} \\
 根据题目提供的条件，我们一步一步推理：\\
 (According to the information in the question, we analyze step by step) \\
  分析每个动物的特征：\\
  (Analyze the characteristics of animals)\\
  ... [text omitted]...\\
  桂鱼：没有腿，没有壳，\textcolor{darkgreen}{生活在淡水。}\\
  (Mandarin fish: no legs, no shell, \textcolor{darkgreen}{lives in freshwater.})\\
  ... [text omitted]...\\
  答案：【B】 \textcolor{darkgreen}{\textbf{√}}\\
  (Answer:[B])\\
\textbf{\colorbox{gray!30}{Model Answer-EN}} \\
 Let's analyze the given information step by step.\\
  Animals and Their Characteristics:\\
  ... [text omitted]...\\
  Mandarin Fish   - Legs: 0   - Shell: No   \textcolor{red}{- Habitat: Lives in seawater}\\
  ... [text omitted]...\\
  Answer: [D]  \textcolor{red}{×}\\
\end{mybox}
\caption{Example of entity property error in the model's response.}
\label{fig:entity_frequency}
\end{figure}

Some entities are related to geography or culture, which can result in their varying frequency across different language corpora. For example, the entity "mandarin fish" is a kind of freshwater fish primarily found in East Asia and North Asia. As a result, it appears more frequently in Chinese corpora. Although the overall size of English data is larger, the data related to the mandarin fish is more abundant in Chinese. The model performs better in understanding commonsense knowledge related to such entities in the Chinese than in English.
 As shown in Figure \ref{fig:entity_frequency}, the model can correctly identify the mandarin fish as a freshwater fish in Chinese but mistakenly classified it as a saltwater fish in English, leading to subsequent reasoning errors.
 
 Similar cases occurs with other entities that are common in Chinese but not so in English, such as "panax notoginseng" (usually used as traditional Chinese medicine) or "winter bamboo shoots" (usually eaten by Asian people), etc.

 \paragraph{Difference in semantic ambiguity}

 \begin{figure}[h!]
\centering
\begin{mybox}[title={}]
\tiny 
\fontsize{7}{5}\selectfont \textbf{\colorbox{gray!30}
{Question-CN}} \\
一面墙上贴着桃子、啤酒、吸管(straw, a tube used to suck liquids into the mouth)、板栗四种物品的照片。已知：\\2号照片上的物品的可食用部分是它的种子；\\4号照片上中的物品不属于工具；\\1号照片上的物品的可食用部分是它的果实；\\2号照片上的物品不属于饮品；\\3号照片上的物品不属于坚果；\\4号照片上中的物品不属于水果；\\3号照片上的物品不属于水果；\\4号照片上中的物品不属于坚果。\\
植物在\_\_\_\_号照片上。\\
\textbf{\colorbox{gray!30}{Options-CN}} \\
A. 1 \\
B. 2 \\
C. 3 \\
D. 4 \\
\textbf{\colorbox{gray!30}{Question-EN}} \\
On a wall pasted photos of four different items: peach, malt liquor, straw, chestnut. Now we know that: \\The edible part of the item on photo No.2 is its seed; \\The item on photo No.4 is not a tool; \\The edible part of the item on photo No.1 is its fruit; \\The item on photo No.2 is not a drink; \\The item on photo No.3 is not a nut; \\The item on photo No.4 is not a fruit; \\The item on photo No.3 is not a fruit; \\The item on photo No.4 is not a nut.\\
A plant is on photo No.\_\_\_\_.\\
\textbf{\colorbox{gray!30}{Options-EN}}\\
A. 1 \\
B. 2 \\
C. 3 \\
D. 4 \\
\textbf{\colorbox{gray!30}{Correct Answer}} \\
AB \\
\textbf{\colorbox{gray!30}{Model Name}} \\
o1-mini \\
\textbf{\colorbox{gray!30}{Model Answer-CN}} \\
 ... [text omitted]...\\
照片1：可食用部分是果实，符合桃子的特性。\\
(Photo 1: edible part is its fruit, peach fits this description.) \\
照片2：可食用部分是种子，并且不属于饮品，符合板栗的特性。\\
(The edible part is its seed, chestnut fits this description.)\\
照片3：不属于坚果和水果，符合吸管的特性。\\
(Not a nut or a fruit, straw fits this description.)\\
照片4：不属于工具、水果和坚果，符合啤酒的特性。\\
(Not a tool, fruit, or nut, malt liquor fits this description.)\\
关于“植物在\_\_\_\_号照片上”，桃子和板栗都是植物的部分（桃树和板栗树的果实和种子），因此植物对应的是照片1和照片2。\\
(Regarding "A plant is on photo No.\_\_\_\_," both peach and chestnut are parts of plants (the fruits and seeds of peach trees and chestnut trees), so the plant corresponds to photo 1 and photo 2.)\\
答案是：【A, B】 \textcolor{darkgreen}{\textbf{√}}\\
  (Answer:[A, B])\\
\textbf{\colorbox{gray!30}{Model Answer-EN}} \\
  ... [text omitted]...\\
  - Photo 2: The edible part is its seed. Among the options, chestnut fits this description.\\
  - Photo 1: The edible part is its fruit. Peach is the fruit here.\\
  - Photo 4: Not a tool, fruit, or nut, so it must be malt liquor.\\
  - Photo 3: Not a nut or a fruit, so it must be straw.\\
  Identifying the Plant: \\ 
  - Straw is derived directly from plants, specifically from cereal grains. \\
  Therefore, the plant is on Photo No.3.\\
  Answer: [C]  \textcolor{red}{×}\\
\end{mybox}
\caption{Example of entity property error in the model's response.}
\label{fig:semantic_ambiguity}
\end{figure}
A same word can have multiple different senses. This semantic ambiguity is more common than in English than in Chinese.
 A same English word may correspond to related or completely unrelated Chinese words, which may bring ambiguity to English questions. 
 The model may confuse different senses during the reasoning process, using different senses in different steps.
 For example, the English word "straw" can correspond to "吸管"(a tube used to suck liquids into the mouth) and "稻草"(thin, hollow stem of grains after harvest) in Chinese. As shown in Figure~\ref{fig:semantic_ambiguity}, in the Chinese question, the word "吸管" has no ambiguity and refers to "a tube used to suck liquids into the mouth", which is a tool. In the English question, the word "straw" is ambiguous. When analyzing the item on photo 4, the model excludes the straw for the item on photo 4 is not a tool, which seems that it uses the sense "a tube used to suck liquids into the mouth". However, when identifying the plant, it uses the sense "thin, hollow stem of grains after harvest" and catagorized the straw as a plant.

Similar cases occurs on other ambiguous English entity names that can refer to different entities, such as "seal" (marine mammal, stamp affixed to a document, or fastener used to securely close something), "fish" (marine animal or meat of fish), etc.



\newpage
\section{Scenario Diagram}
\label{app:scenario_diagram}
\begin{figure*}[h]
  \centering
  \includegraphics[width=1\textwidth]{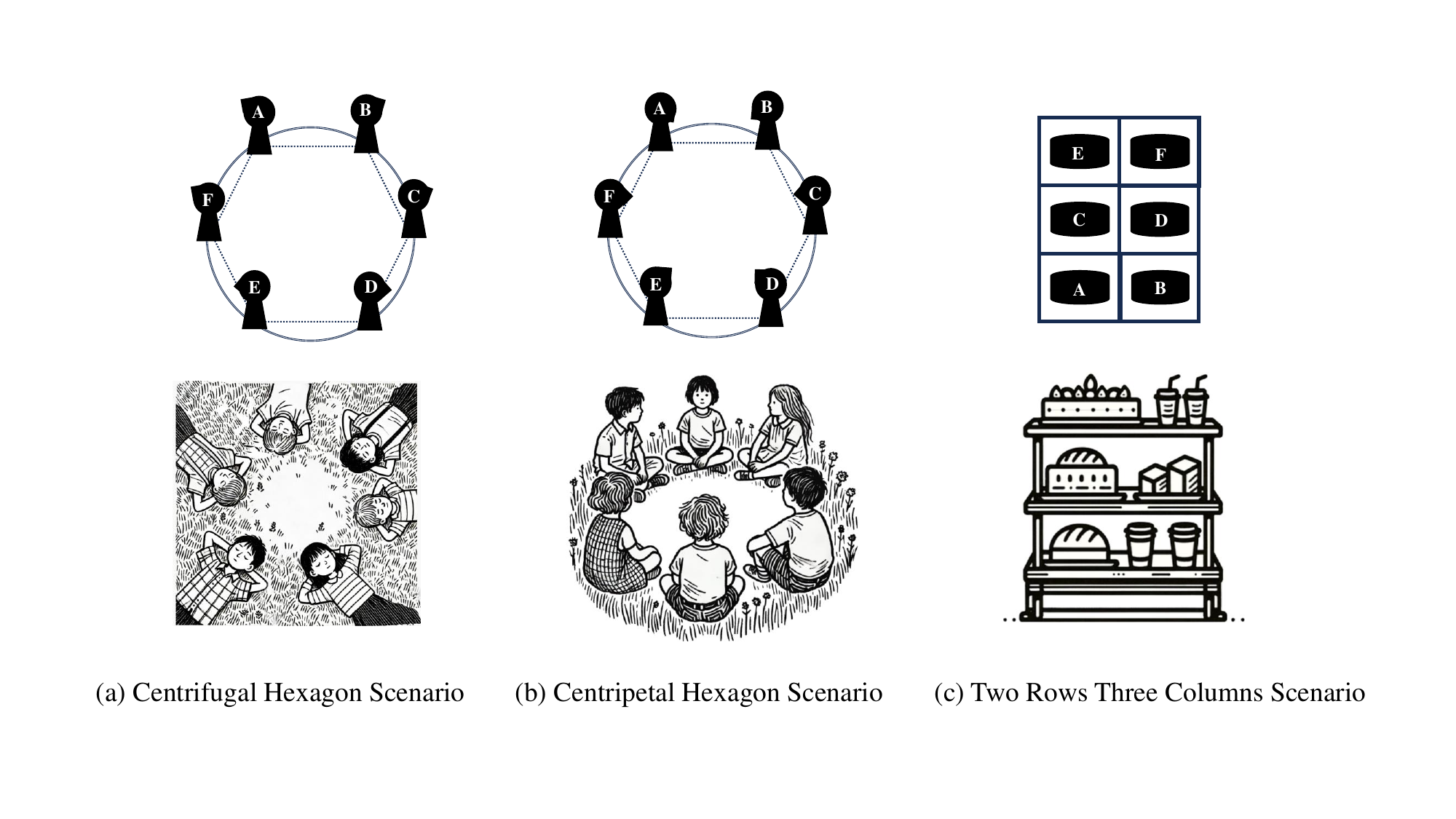}
  \caption {Spatial Scenario Diagram}
  \label{fig:space-scenario}
\end{figure*}

The space scenario diagram is shown as Figure \ref{fig:space-scenario}.

\begin{figure*}[h]
  \centering
  \includegraphics[width=1\textwidth]{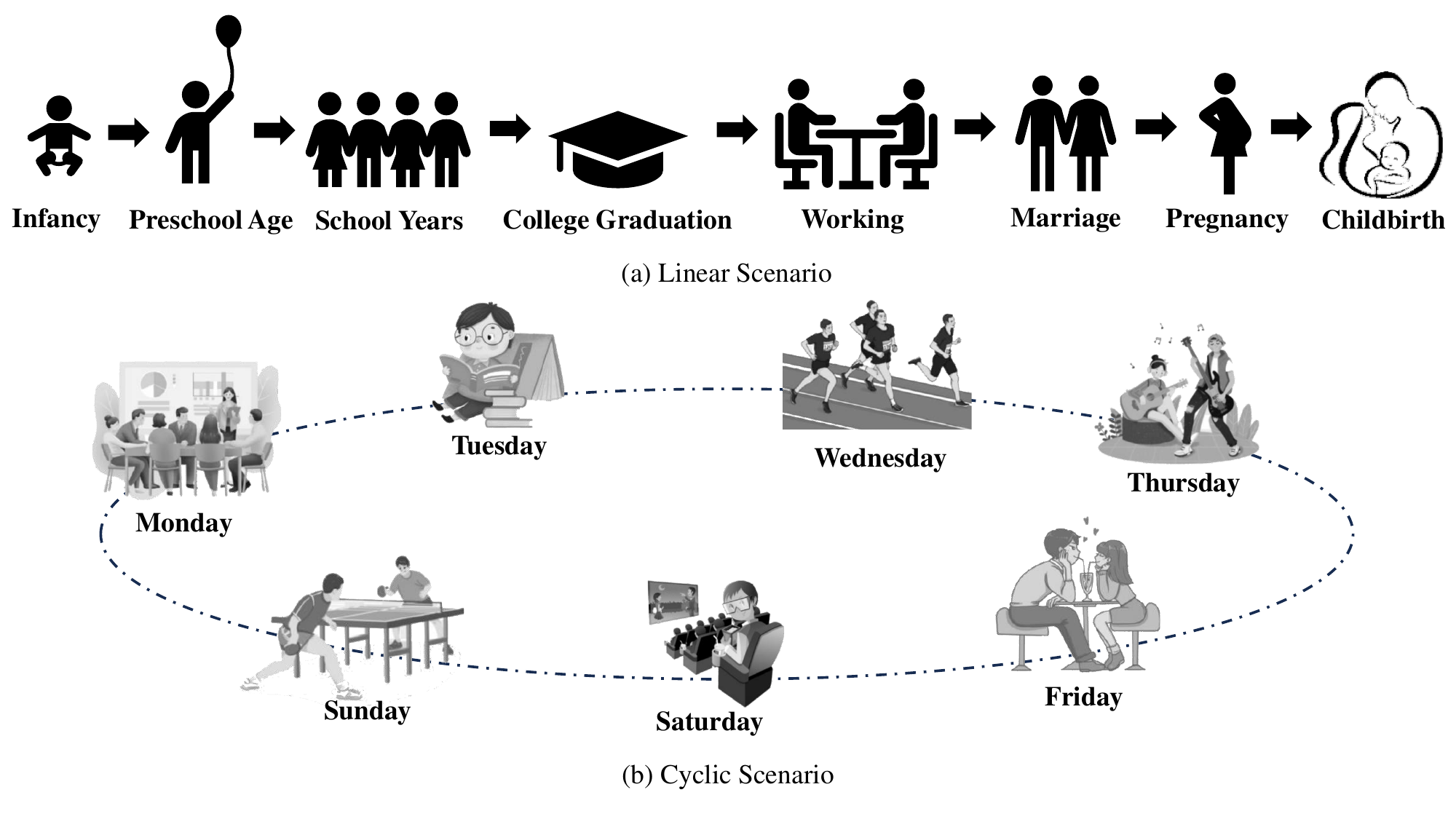}
  \caption {Time Scenario Diagram}
  \label{fig:time-scenario}
\end{figure*}

\noindent The time scenario diagram is shown as Figure \ref{fig:time-scenario}.

\begin{figure*}[h]
  \centering
  \includegraphics[width=1\textwidth]{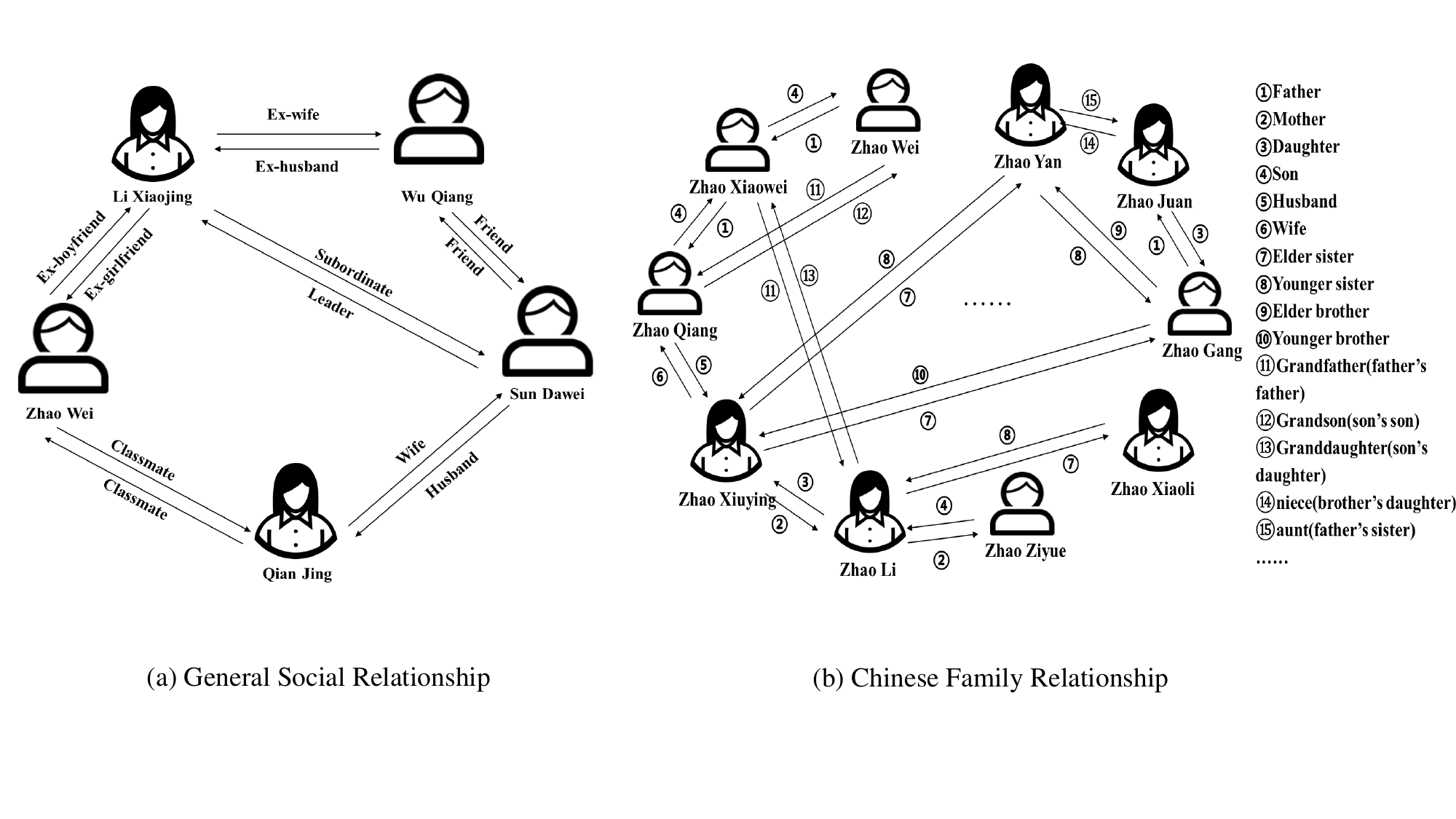}
  \caption {Social Scenario Diagram}
  \label{fig:social-scenario}
\end{figure*}

\noindent The social scenario diagram is shown as Figure \ref{fig:social-scenario}.

\begin{figure*}[h]
  \centering
  \includegraphics[width=1\textwidth]{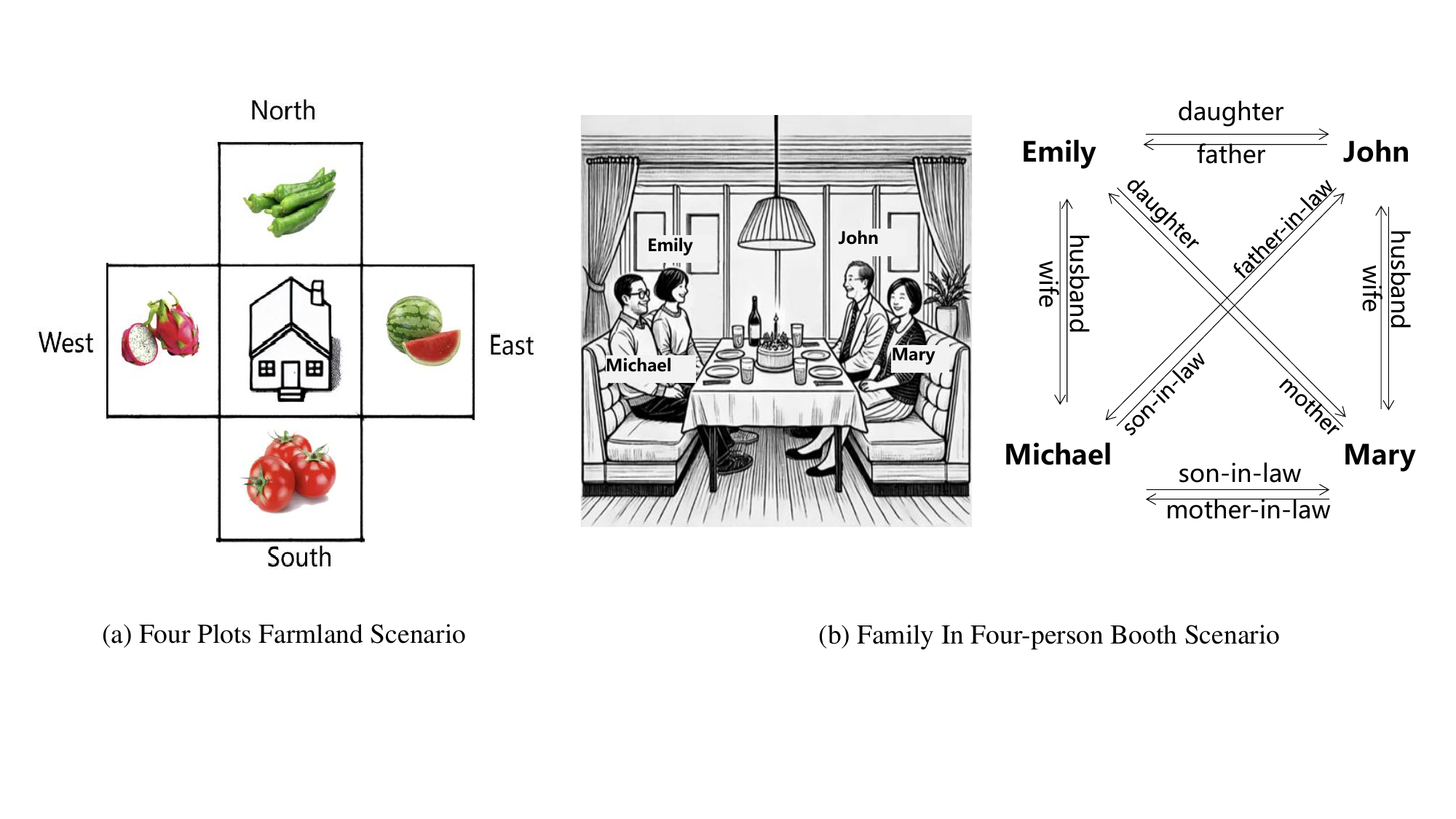}
  \caption {Enhanced Spatial Scenario Diagram with Entity Properties and Relationships}
  \label{fig:enhanced-scenario}
\end{figure*}

\noindent The enhanced spatial scenario diagram is shown as Figure \ref{fig:enhanced-scenario}.

\end{CJK*}

\end{document}